%% file: paper.tex
\documentclass[]{memtensor}
\usepackage{enumitem}
\usepackage[toc,page,header]{appendix}
\usepackage[utf8]{inputenc}
\usepackage[T1]{fontenc}
\usepackage{hyperref}
\usepackage{url}
\usepackage{graphicx}
\usepackage{booktabs}
\usepackage{amsfonts}
\usepackage{nicefrac}
\usepackage{makecell}
\usepackage{microtype}
\usepackage{amsmath}
\usepackage{etoolbox}
\usepackage{lipsum}  
\usepackage{minitoc}
\usepackage{tablefootnote}
\usepackage{threeparttable}
\usepackage{wrapfig}
\usepackage{appendix}
\usepackage{multirow}
\usepackage{ulem}
\useunder{\uline}{\ul}{}
\usepackage{colortbl}

\title{
  \raisebox{-0.3\height}{\includegraphics[width=0.08\linewidth]{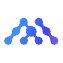}}
  \titlefont \textsc{MemOS}: A Memory OS for AI System}

\author[1,2]{Zhiyu Li}
\author[1]{Chenyang Xi}
\author[1]{Chunyu Li}
\author[3]{Ding Chen}
\author[1]{Boyu Chen}
\author[1,8]{Shichao Song}
\author[1,8]{Simin Niu}
\author[1,8]{Hanyu Wang} 
\author[1,8]{Jiawei Yang}
\author[1]{Chen Tang}
\author[1,9]{Qingchen Yu}
\author[1,8]{Jihao Zhao}
\author[1]{Yezhaohui Wang} 
\author[8]{Peng Liu}
\author[1]{Zehao Lin}
\author[1]{Pengyuan Wang}
\author[1]{Jiahao Huo}
\author[1,10]{Tianyi Chen} 
\author[1,2]{Kai Chen}
\author[1,3]{Kehang Li}
\author[8]{Zhen Tao}
\author[1]{Huayi Lai}
\author[1]{Hao Wu}
\author[1]{Bo Tang}
\author[7,2]{Zhengren Wang}
\author[9]{Zhaoxin Fan}
\author[5]{Ningyu Zhang}
\author[10]{Linfeng Zhang}
\author[10]{Junchi Yan}
\author[3]{Mingchuan Yang}
\author[6]{Tong Xu}
\author[8]{Wei Xu}
\author[5]{Huajun Chen}
\author[4]{Haofen Wang}
\author[1,2]{Hongkang Yang}
\author[7,\dag]{Wentao Zhang}
\author[10,\dag]{Zhi-Qin John Xu}
\author[10,\dag]{Siheng Chen}
\author[1,2,\dag]{Feiyu Xiong}

\affiliation[1]{MemTensor (Shanghai) Technology Co., Ltd.}
\affiliation[2]{Institute for Advanced Algorithms Research, Shanghai}
\affiliation[3]{Research Institute of China Telecom}
\affiliation[4]{Tongji University}
\affiliation[5]{Zhejiang University}
\affiliation[6]{University of Science and Technology of China}
\affiliation[7]{Peking University}
\affiliation[8]{Renmin University of China}
\affiliation[9]{Beihang University}
\affiliation[10]{Shanghai Jiao Tong University}

\abstract{
Large Language Models (LLMs) have become an essential infrastructure for Artificial General Intelligence (AGI), yet their lack of well-defined memory management systems hinders the development of long-context reasoning, continual personalization, and knowledge consistency.Existing models mainly rely on static parameters and short-lived contextual states, limiting their ability to track user preferences or update knowledge over extended periods.While Retrieval-Augmented Generation (RAG) introduces external knowledge in plain text, it remains a stateless workaround without lifecycle control or integration with persistent representations.Recent work has modeled the training and inference cost of LLMs from a memory hierarchy perspective, showing that introducing an explicit memory layer between parameter memory and external retrieval can substantially reduce these costs by externalizing specific knowledge \cite{memory3_Yang_2024}. Beyond computational efficiency, LLMs face broader challenges arising from how information is distributed over time and context, requiring systems capable of managing heterogeneous knowledge spanning different temporal scales and sources. To address this challenge, we propose \textsc{MemOS}, a memory operating system that treats memory as a manageable system resource. It unifies the representation, scheduling, and evolution of plaintext, activation-based, and parameter-level memories, enabling cost-efficient storage and retrieval. As the basic unit, a MemCube encapsulates both memory content and metadata such as provenance and versioning. MemCubes can be composed, migrated, and fused over time, enabling flexible transitions between memory types and bridging retrieval with parameter-based learning. \textsc{MemOS} establishes a memory-centric system framework that brings controllability, plasticity, and evolvability to LLMs, laying the foundation for continual learning and personalized modeling.
}

\checkdata[Author Legend]{\dag{}Correspondence}
\checkdata[Project Website]{\url{https://memos.openmem.net/}}
\checkdata[Code]{\url{https://github.com/MemTensor/MemOS}}
\begin{document}
\maketitle

\newpage

\section{Introduction}

\begin{figure}[ht]
    \centering
    \includegraphics[width=0.8\linewidth]{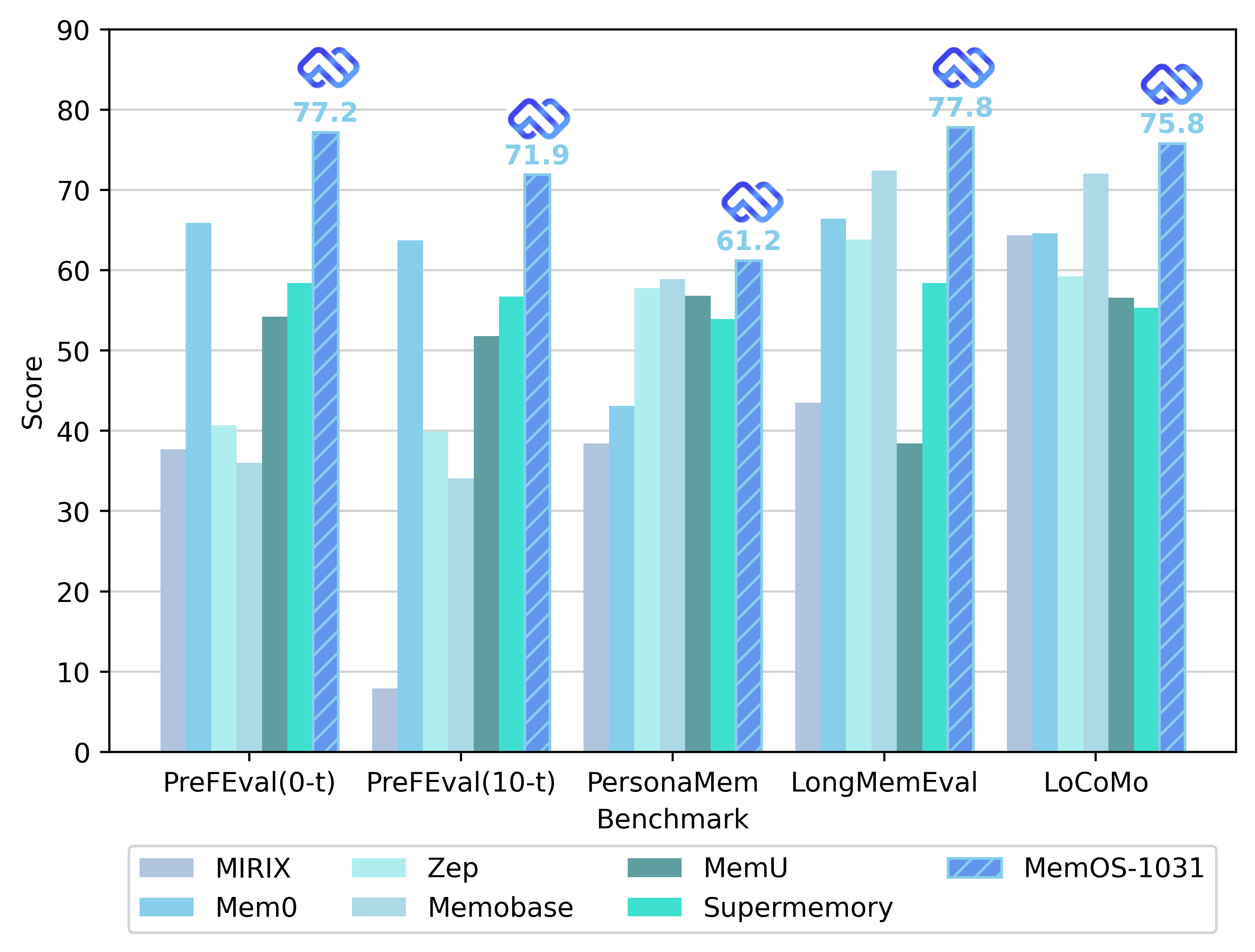}
    \caption{
    \textbf{\textsc{MemOS} achieves state-of-the-art performance across all benchmarks.} This figure provides a comprehensive summary of our evaluation results, illustrating the personalized response rate on PreFEval(0 turns and 10 turns), the precision score on PersonaMem, the overall mean score on LongMemEval, and the overall mean LLM judge score on the LoCoMo benchmark. The \textsc{MemOS} (MemOS-1031) consistently ranks first in all categories, significantly outperforming strong baselines such as MIRIX, Mem0, Zep, Memobase, MemU, and Supermemory. Full detailed metric breakdowns are provided in Table~\ref{tab:new_locomo}, Table~\ref{tab:perfeval}, Table~\ref{tab:longmemeval} and Table~\ref{tab:personamem}.
    }
    \label{fig:maintest}
\end{figure}

With the advent of the Transformer architecture and the maturation of self-supervised pretraining, Large Language Models (LLMs) have become the cornerstone of modern NLP.
Trained on large-scale corpora, LLMs exhibit near-human performance in open-domain QA, text generation, and summarization tasks \cite{zhao2023survey}.
With increasing model size and compute, their capabilities have expanded to structured code generation \cite{wang2023codet5}, cross-modal reasoning \cite{qian2024linguistic}, multi-turn dialogue, and complex planning—positioning LLMs as a leading paradigm toward Artificial General Intelligence (AGI).

Looking ahead, the presence of LLMs, or more generally, AGI systems, will expand vastly in both time and space.
Temporally, models will shift from stateless, session-based tools to persistent agents embedded in long-running workflows.
Much like humans, they will need to accumulate interaction histories, adapt internal states, and reason over extended contexts.
Spatially, LLMs are evolving into foundational intelligence layers across users, platforms, and ecosystems.
Whether deployed in cloud services or embedded in enterprise systems, they must support consistency, adaptability, and personalization across users, roles, and tasks.
As such omnipresence becomes the norm, a critical challenge emerges: how should knowledge be organized, stored, and retrieved?

With expanding interaction histories, models face a potentially unbounded context space.
We anticipate that future LLMs will seek to leverage as much of their accessible temporal and spatial context as possible, to support deeper reasoning, decision-making, and adaptation.
No longer reprocessing all past information per inference, they will decide what to retain, compress, discard, or prioritize.
In this always-on paradigm, memory becomes a necessity, not an add-on, for maintaining coherent behavior and identity over time.
This requires efficient management of large-scale, multi-source information and dynamic scheduling of memory conditioned on context.
This motivates a layered memory hierarchy, similar to how OSs manage memory, consisting of working memory, long-term storage, and cold archives, governed by recency, access frequency, and importance.
Sharing memory across users and agents requires scoping, permission control, and migratable, reusable representations.
These capabilities are vital not only for system efficiency, but for the long-term sustainability of model-based knowledge evolution.

The management of memory will become model-defined instead of human-defined.
Just as deep learning replaced feature engineering, the transition of memory management from hard-coded pipelines (e.g., RAG) to learnable strategies is natural and necessary.
Future agents will autonomously decide whether to retrieve memory, summarize interaction into reusable rules, abstract preferences, or transfer knowledge across contexts.
In essence, models must take on the responsibility of shaping their own memory architectures and strategies.
Yet, existing infrastructures fall short of enabling this shift.

Mainstream LLMs rely on implicit \textbf{parameter memory}, encoding knowledge in billions or trillions of model weights.
While this approach affords generalization, it suffers from high update cost, poor interpretability, and limited flexibility.
Retraining or fine-tuning requires significant computational resources and risks issues such as catastrophic forgetting.

To address this bottleneck, Retrieval-Augmented Generation (RAG) has emerged as a popular augmentation strategy.
By incorporating external retrieval modules, RAG allows models to dynamically access fresh information at inference time, enabling augmentation without parameter updates \cite{DBLP:journals/corr/abs-2402-19473, DBLP:journals/corr/abs-2312-10997, DBLP:journals/corr/abs-2501-13958, DBLP:journals/corr/abs-2502-06872, DBLP:journals/corr/abs-2310-05029, DBLP:journals/corr/abs-2404-16130, DBLP:journals/corr/abs-2410-05779}.
It is now widely deployed in systems such as Copilots \cite{rag_for_copilots} and enterprise search \cite{vertex_AI_search, build_innovative_AI_search_experiences, Agentic_RAG_as_a_Service_company, vu_freshllms_arxiv23}.
Nonetheless, RAG remains fundamentally an ``on-the-fly retrieval and transient composition'' pipeline, rather than an integrated memory management system.
It lacks core memory manageability features such as lifecycle tracking, versioning, and permission-aware scheduling, limiting its ability to support long-term, adaptive knowledge systems.
As a result, models continue to exhibit short-memory behavior in multi-turn dialogue, planning, and personalization tasks, struggling to maintain behavioral consistency or long-horizon adaptation.

\begin{figure}[ht]
    \centering
    \includegraphics[width=0.55\textwidth]{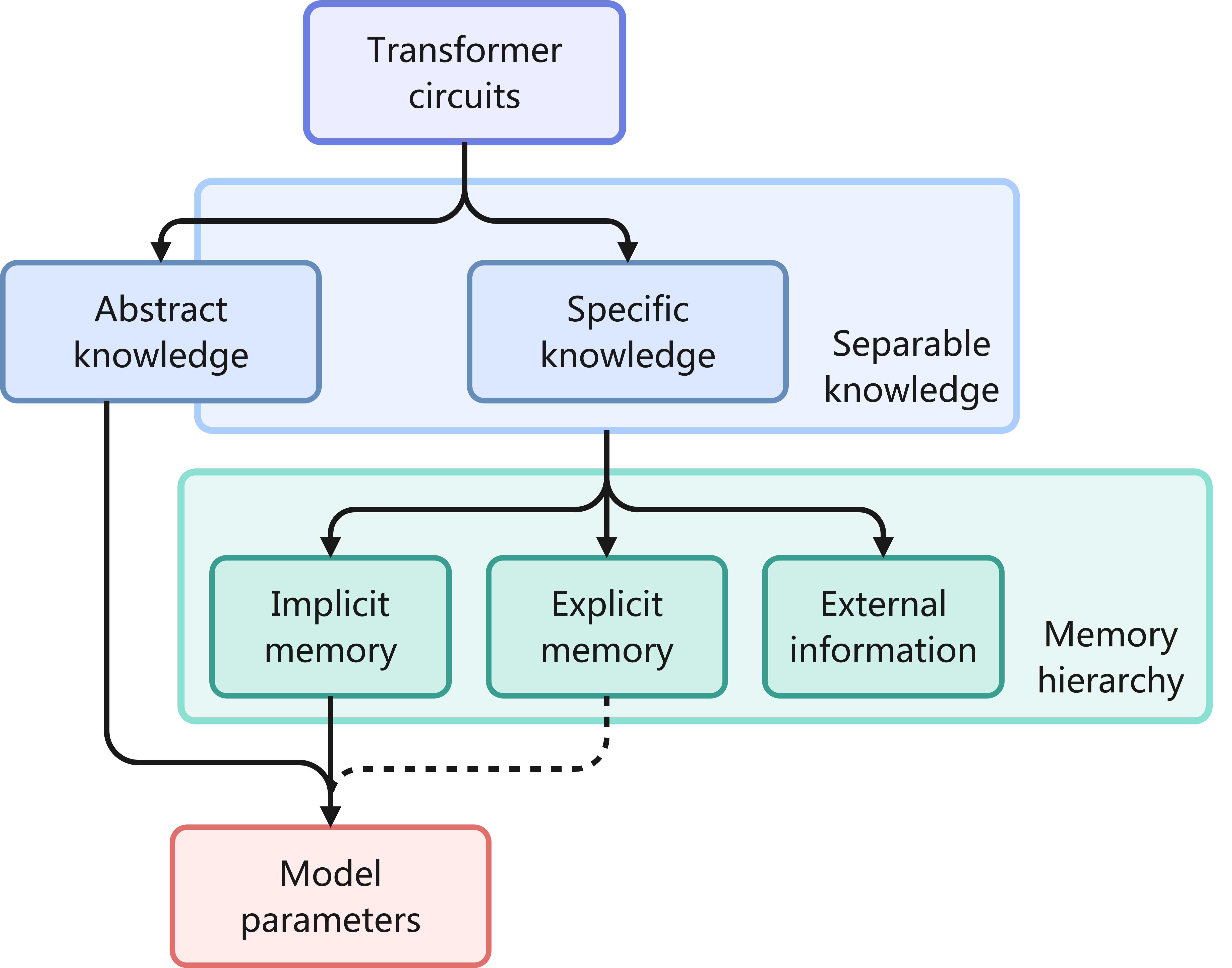}
    \caption{
        Categorization of LLM knowledge, including the memory hierarchy.
        The explicit memories, extracted from model activations, lie half-way between raw data and model parameters,
        so a dotted line is used to indicate that they may or may not be regarded as parameters.
        Reproduced from~\cite{memory3_Yang_2024}.
    }
    \label{fig:memory_categorization}
\end{figure}

Recent work has shown that the limitations of current memory mechanisms are not incidental, but stem from the architectural absence of explicit and hierarchical memory representations within LLMs.
For example, \cite{memory3_Yang_2024} argues that without an intermediate explicit memory layer bridging external retrieval and parametric storage, models become suboptimal in terms of read-write cost, and cannot balance storage cost against retrieval efficiency.
This distinction is illustrated in Figure~\ref{fig:memory_categorization}, which categorizes knowledge and memory formats and highlights the intermediate role of explicit memory.

From a systems perspective, neither parametric memory nor RAG treats memory as a schedulable and evolvable system resource.
This structural gap remains a core bottleneck preventing LLMs from becoming persistent and collaborative intelligent agents.
As application scenarios grow more complex, these limitations become particularly evident in the following four typical contexts.

\begin{itemize}
    \item \textbf{Long-range Dependency Modeling}: As tasks and dialogues grow in length, models must preserve instruction and state consistency across multiple turns or stages.
    However, current Transformer architectures face three major obstacles: limited context windows constrain input capacity, quadratic attention cost leads to high compute overhead, and user instructions often detach from model behavior over long horizons.
    For example, in complex tasks, user-defined code structures or writing styles are frequently forgotten, and model outputs revert to default modes.
    As LLMs are deployed in multi-turn dialogue, long-form generation, and persistent workflows, long-context—and even infinite-context—will become a general requirement rather than a rare exception.
    This limitation indicates the lack of mechanisms for persistent state maintenance and structured context retention.

    \item \textbf{Adapting to Knowledge Evolution}: Real-world knowledge evolves continuously (e.g., legal updates, scientific discoveries, current events), but static parameters prevent timely reflection. RAG allows dynamic retrieval, yet remains a stateless patching mechanism lacking unified versioning, provenance, or temporal awareness. For instance, it may cite outdated and new regulations simultaneously without reconciliation. It cannot retire obsolete facts, prioritize reliable ones, or track knowledge evolution—limiting long-term consistency.

    \item \textbf{Personalization and Multi-role Support}: LLMs lack durable “memory traces” across users, roles, or tasks. Each session resets to a blank state, ignoring accumulated preferences or styles. Although tools like ChatGPT and Claude now offer memory, issues persist: capacity limits, unstable access, opaque updates, and missing editability. Current systems emphasize passive recording over structured control, making them ill-suited for long-term personalization across diverse use cases.

    \item \textbf{Cross-platform Memory Migration and Ecosystem Diversity}: As LLMs expand from single interfaces to multi-end deployments (web, mobile, enterprise), user memories (e.g., profiles, task history, preferences) should persist across contexts. Yet most systems trap memory within specific instances, forming “memory islands.” For example, ideas explored in ChatGPT \cite{achiam_2023_gpt4} can't carry over to Cursor \cite{noauthor_cursor_nodate}, forcing context rebuilding. This impairs continuity and blocks memory reuse. Deeper yet, centralization vs. decentralization poses a systemic challenge: while monopolized platforms benefit from feedback loops, distributed models risk stagnation. Making memory portable and reusable is key to balancing evolution efficiency with ecosystem diversity.
    
\end{itemize}

A review of the four challenges reveals a shared pattern: models lack the ability to coherently manage and coordinate information distributed across time and space.
This is not due to any single failing module, but to the absence of a system-level mechanism for organizing and operating over memory.

Modern LLMs lack an intermediate layer between parametric storage and external retrieval, making it difficult to manage memory lifecycle, integrate evolving knowledge, or maintain behavioral continuity.
While RAG provides access to external information, its lack of unified structure and operational semantics prevents long-term, controllable use of knowledge.

Therefore, we argue that building future-capable language intelligence systems requires treating memory as a system-level resource that can be explicitly modeled and scheduled.
In modern operating systems, computational resources (CPU), storage (RAM/disks), and communication (I/O) are uniformly scheduled and managed across their lifecycle.
In contrast, memory in large model architectures exists as implicit parameters or temporary retrievals—neither schedulable nor traceable, and incapable of integration or transfer.
Therefore, the key to enhancing memory in LLMs is not simply ``adding a cache" or ``attaching an external retrieval module," but redefining the operational logic and resource management of memory from a systems-level perspective.

To address these challenges, we propose \textsc{MemOS} (Memory Operating System), a dedicated memory operating system designed for large language models.
The core philosophy of \textsc{MemOS} is that, in order to fully utilize temporally and spatially distributed information, models require a unified framework for organizing memory, maintaining internal state, and supporting long-term adaptation.

Inspired by recent work on memory hierarchy for improving model efficiency and adaptability \cite{memory3_Yang_2024}, \textsc{MemOS} extends this idea into a system-level design by modeling memory as schedulable and evolvable resource units.
It builds a modular architecture around the memory lifecycle—including generation, activation, fusion, archiving, and expiration—supported by components such as \texttt{MemReader}, \texttt{MemScheduler}, \texttt{MemLifecycle}, and \texttt{MemOperator}, which together orchestrate memory flow, state transitions, and access control.

Much like traditional operating systems coordinate CPU, memory, and I/O, \textsc{MemOS} provides an abstraction layer and unified \texttt{Memory API}, enabling consistent and auditable access to memory units across users, tasks, and sessions.
The system supports structured storage, provenance tagging, lifecycle tracking, and fine-grained permission enforcement, forming a scalable foundation for memory-driven reasoning.
More importantly, \textsc{MemOS} lays a cognitive foundation for the next generation of AGI systems with long-term memory and continual evolution, and provides efficient infrastructure for memory-centric architectural innovation.

The system provides three core capabilities:

\begin{itemize}

    \item \textbf{Controllability}: \textsc{MemOS} offers full lifecycle management of memory units, enabling unified scheduling of memory creation, activation, fusion, and disposal. It implements multi-level permission control and context-aware activation strategies, ensuring safety and traceability in multi-task and multi-user environments through access control and operation auditing. For instance, user preference memories can be scoped to specific agent instances and automatically expire or archive after task completion.

    \item \textbf{Plasticity}: \textsc{MemOS} supports memory restructuring and migration across tasks and roles. It provides memory slicing, tagging, hierarchical mapping, and context binding capabilities, allowing developers or systems to construct highly adaptable memory structures based on inference objectives. This enables models to activate different memory views for different tasks or update memory associations dynamically during role transitions, facilitating rapid cognitive adaptation and behavior shaping.

    \item \textbf{Evolvability}: \textsc{MemOS} enables dynamic transitions and unified scheduling among different memory types—including parameter memory (knowledge embedded in model weights), activation memory (contextual inference state), and plaintext memory (structured knowledge fragments). The system supports seamless transitions, such as converting user-defined rules from multiple dialogues into active memory, or compressing long-term structured knowledge into parametric form. This cross-memory adaptation provides a robust foundation for knowledge integration, autonomous learning, and model evolution.

\end{itemize}

Therefore, as a novel infrastructure for the continual evolution of LLMs, \textsc{MemOS} aims to reconstruct the representation, management, and scheduling of memory from a systems perspective. It addresses core limitations in structured memory, lifecycle management, and multi-source integration, while providing OS-level support for cross-task adaptation, cross-modal evolution, and cross-platform migration.
The introduction of \textsc{MemOS} marks a critical transition in the development of large models: from mere perception and generation to memory and evolution.

\section{Memory in Large Language Models}
Research in memory capabilities in large language models has generally progressed through four key stages:
\textbf{(1) The stage of definition and exploration}, which focuses on categorizing and analyzing LLM memory systems from multiple perspectives, while identifying effective optimization mechanisms applicable in real-world scenarios.
\textbf{(2) The stage of human-like memory development}, which addresses performance gaps in complex tasks arising from discrepancies between LLM and human memory by introducing various forms of cognitively inspired memory mechanisms.
\textbf{(3) The stage of tool-based memory management}, where modular interfaces for memory operations begin to emerge, yet are largely limited to basic insert, delete, and update functionalities over existing memory structures.
Our proposed \textsc{MemOS} introduces operating system–inspired resource management mechanisms to LLM memory, offering standardized and unified interfaces for full-lifecycle memory management and scheduling. This paves the way toward
\textbf{(4) The stage of systematic memory governance}, enabling structured evolution, abstraction, and secure control over memory resources.
In this subsection, we review existing research on memory in large models along this developmental trajectory.

\begin{figure}[htp]
    \centering
    \includegraphics[width=1.0\linewidth]{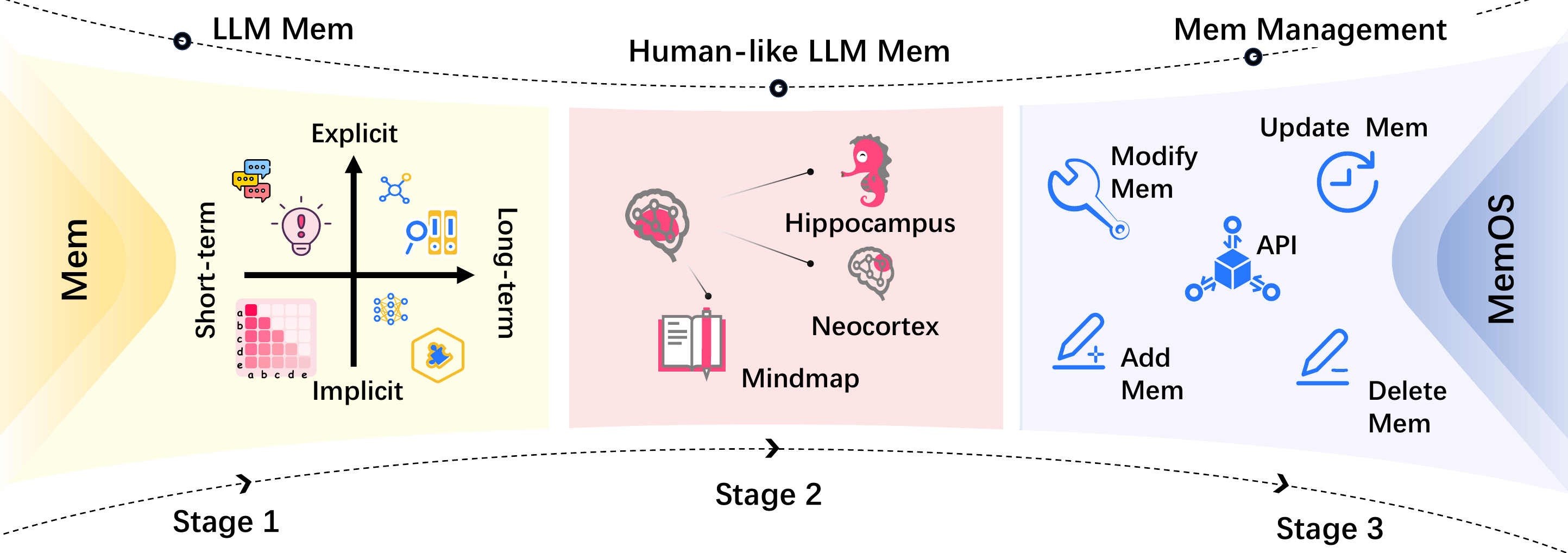}
    \caption{Illustration of the evolution of memory systems in large language models, highlighting the progression from definition and exploration, to human-like memory development, and to tool-based memory management.}
    \label{fig:Memory in LLMs}
\end{figure}

\subsection{Stage 1: Memory Definition and Exploration}
Several recent studies have proposed systematic classifications and analyses of memory in LLMs from various dimensions. For example, \cite{du_cuhkSurvey_2025} categorizes memory into three types: parameter memory, unstructured contextual memory, and structured contextual memory. \cite{wu_huaweiSurvey_2025} classifies memory based on object (personal vs. system), form (parametric vs. non-parametric), and temporal aspects (short-term vs. long-term). \cite{shan_LiAutoSurvey_2025} further divides memory into four types: parameter-based, key-value cache-based, hidden state-based, and text-based, and introduces retention duration as a standard to distinguish sensory memory, short-term memory, and long-term memory.

Building on these works, we propose that LLM memory can be characterized along two primary dimensions:  \textbf{implicit} and 	\textbf{explicit}. Implicit memory includes parameter memory, key-value cache, and hidden states, while explicit memory involves text- and context-based information storage.
Memory can be classified temporally as sensory, short-term, or long-term. Sensory memory captures fleeting impressions of perceptual input, with extremely short duration and no conscious processing. While traditionally treated as a separate stage, we include it under short-term memory for unified scheduling and handling of initial information. This work adopts this two-dimensional framework to analyze memory mechanisms in the first and second stages (see Figure \ref{fig:Memory in LLMs} left, Table \ref{tab:memory_classification}).

\input{tab/references}

\subsubsection{Implicit Memory in LLMs}
\paragraph{\textbf{Implicit Long-term Memory in LLMs}}
Through large-scale pretraining, LLMs encode syntactic structures, conceptual relationships, and language usage patterns from corpora into their weight matrices. These parameters serve as implicit long-term memory, internalized into the model's inherent capabilities. Although they lack explicit expression, they continuously influence language generation behavior, knowledge expression, and even semantic generalization.

\textbf{Training:}
In LLMs, training is the most fundamental and direct method for forming implicit long-term memory. For example, pretraining~\cite{devlin_bert_2019,brown_gpt3_2020} and post-training~\cite{bai_rlhf_2022,ouyang_instructgpt_2022} enable large-scale parameter updates, fundamentally reconstructing the internal knowledge distribution and behavioral structure of the model.
Some studies introduce memory explicitly during training. For instance, CTRL~\cite{keskar_ctrl_2019} includes control codes in training data to help models automatically associate contextual information during text generation. Memory\&Reasoning~\cite{jin_MemandReason_2024} fine-tunes the model to decouple output into separate memory and reasoning components, fully leveraging memory for inference. SLayer~\cite{chen_slayer_2024} identifies memory-relevant layers in the model and locally fine-tunes them to enhance specific knowledge representation.
It is worth noting that relying solely on memorization of training data can be limited in real-world deployment due to distributional shifts between real-world and training data.
Titans~\cite{DBLP:journals/corr/abs-2501-00663} proposes a dynamic memory mechanism by encoding historical information into neural network parameters and training a pluggable online meta-model. This meta-model can adaptively decide retention or forgetting strategies for specific data during real usage, thereby improving generalization across distribution shifts.

\textbf{Adaptor:}
Full-scale training or fine-tuning is costly and often impractical for rapid memory updates in real-world scenarios.
To address this, adapter-based methods freeze the core model parameters and introduce small, trainable modules that adapt quickly to new memory with minimal disruption to original capabilities.
LoRA~\cite{hu_lora_2021} inserts low-rank adapters into the model, enabling lightweight parameter tuning without modifying the original parameter structure, supporting efficient loading and storage of implicit memory.
PRAG~\cite{su_PRAG_2025} treats LoRA adapter modules trained for specific documents or tasks as “memory units” and merges them into the main model as needed, enabling rapid access to specialized knowledge.
Furthermore, DyPRAG~\cite{tan_DyPRAG_2025} introduces a neural generator that directly maps input documents to LoRA parameters, significantly reducing explicit memory storage cost.

\textbf{Editing:}
Memory editing refers to targeted interventions on model parameters to induce new knowledge or behaviors for specific inputs while preserving existing capabilities as much as possible.
Most existing research focuses on editing objective factual knowledge, such as correcting answers to questions like “Who is the president of the United States?”
However, memory in LLMs also includes abstract competencies such as language style, semantic preferences, and reasoning modes, for which systematic editing methods are still lacking.
If not carefully controlled, local parameter edits can lead to undesirable global behavior shifts. Thus, edit precision and retention of existing capabilities are key evaluation metrics.
This paper follows \cite{yao_KESurvey_2023} in categorizing knowledge editing techniques into three types:
\textbf{(1) Locate-then-edit intuitive methods}\cite{meng_rome_2023,meng_memit_2023,xu_biasedit_2025}: These methods use causal tracing to locate where the target knowledge is stored, followed by targeted parameter updates.
\textbf{(2) Meta-learning-based methods}\cite{cao_KnowledgeEditor_2021,mitchell_mend_2021,tan_malmen_2023}: These use hypernetworks to directly predict parameter changes. Another important direction is preserving prior knowledge and abilities during editing\cite{fang_alphaedit_2025,jiang_anyedit_2025,li_AdaPLE_2024}.
\textbf{(3) Adapter-based editing strategies}\cite{dong_CaliNet_2022,mitchell_serac_2022,hartvigsen_grace_2023,cheng_DPM_2023}: These preserve the LLM backbone, offering a degree of edit controllability.

\paragraph{\textbf{Implicit Short-term Memory in LLMs}}

Beyond the internalized parametric long-term memory, LLMs also depend on dynamically generated and transient intermediate representations during inference—such as KV-caches and hidden states. Although these representations lack explicit forms, they continually influence attention distributions and behavioral strategies in autoregressive generation, forming the implicit short-term memory of LLMs. They play a vital role in maintaining contextual coherence, enabling instant control, and facilitating behavior transition, and have become a crucial entry point for understanding and enhancing dynamic capabilities of language models.

\textbf{KV-cache:} 
KV-cache stores key-value representations of previously processed tokens, enabling persistent access to historical memory during autoregressive generation. Although users cannot directly manipulate these caches, they implicitly modulate attention and output behavior during inference \cite{memory3_Yang_2024}.
Subsequent optimization work has focused primarily on improving compute and memory efficiency. Techniques such as low-rank compression and quantization are adopted by LESS \cite{dong_less_2024} and KVQuant \cite{hooper_kvquant_2024}, while StreamingLLM \cite{xiao_streamingllm_2023} and H\textsubscript{2}O \cite{zhang_h2o_2023} dynamically prune less relevant KV pairs based on attention patterns.
More recent studies introduce retrieval-based memory activation \cite{tang_quest_2024,liu_retrievalattention_2024}, enabling selective access to cached content. Meanwhile, vLLM \cite{kwon_vLLM_2023} draws from operating system design by implementing PagedAttention—using virtual memory-style page caching to reduce redundant storage and improve KV access.

While most existing work focuses on optimizing KV-cache for inference efficiency, its capacity to represent structured and controllable knowledge remains underexplored. Memory\textsuperscript{3} \cite{memory3_Yang_2024} takes a first step in this direction by encoding external knowledge bases as sparse key-value pairs, which are injected into the model’s self-attention layers. This enables dynamic, non-parametric retrieval of relevant information during inference, effectively externalizing knowledge and improving memory controllability—offering new directions for the structured use of short-term memory.
Building on the foundation laid by Memory\textsuperscript{3}, \textsc{MemOS} advances the notion of structured memory by proposing the first hierarchical memory architecture for LLMs that models and unifies three distinct substrates: plaintext memory, activation memory, and parameter memory. It introduces an integrated retrieval and scheduling framework that enables explicit control, efficient fusion, and dynamic activation. The MemCube module further organizes semantic fragments into a multi-dimensional structure, enabling query-based aggregation and multi-granularity activation—paving the way for more systematic and scalable memory utilization in LLMs.

\textbf{Hidden States:}  
Hidden states represent the layer-wise intermediate activations within LLMs during processing, encoding the model's semantic understanding and generation trajectory. Compared to modifying model parameters, directly manipulating hidden states offers a more flexible, instantaneous, and efficient means of memory control.
Among the various mechanisms, steering vectors \cite{subramani_steerno1_2022} stand out as a representative method. These vectors are derived by computing activation differences between inputs with contrasting semantic attributes, forming directionally meaningful control signals. Injecting such vectors into the intermediate activations of other inputs can steer generation toward specific semantic directions without altering the model architecture.
To avoid reliance on supervised corpora, methods like Self-Detoxifying \cite{leong_selfdetoxifying_2023}, ActAdd \cite{turner_ActAdd_2024}, ICV \cite{liu_ICV_2024}, StyleVec \cite{konen_stylevec_2024}, and CAA \cite{rimsky_CAA_2024} propose unsupervised contrastive approaches. These construct semantically similar yet attribute-opposing input pairs (e.g., emotion, stance, politeness) to extract hidden state differences and generate steering vectors, enabling automated, lightweight signal derivation. This not only enhances the portability of steerable control but also lowers its entry barrier.
As an implicit short-term memory mechanism, hidden states have been validated in various practical tasks. For example, steering vectors have been employed in hallucination mitigation and factual consistency enhancement in ACT \cite{wang_ACT_2025}, ITI \cite{li_ITI_2023}, and InferAligner \cite{wang_inferaligner_2024}. IFS \cite{stolfo_IFS_2025} extends their application to controlling low-level generation features such as text formatting and sentence length, indicating that hidden state interventions are effective not only for abstract semantics but also for structural behavior modulation.

\subsubsection{Explicit Memory in LLMs}
\paragraph{\textbf{Explicit Short-term Memory in LLMs}}
LLMs' explicit short-term memory primarily resides in their input context window—namely, the prompt and directly concatenated historical dialogues, including user task descriptions, interaction history, and reference documents. These explicitly injected elements are directly perceived and utilized during inference, forming the basis for understanding the current context and generating responses.
With the increasing scale and capabilities of LLMs, their ability to manage explicit short-term memory has significantly improved. From early general-purpose language models relying on static text input~\cite{radford_gpt2_nodate,brown_gpt3_2020}, to parameterized prompt techniques using learnable continuous vectors~\cite{li_prefix-tuning_2021,lester_prompt-tuning_2021,liu_ptuning_2023,liu_ptuningv2_2022}, to advanced instruction-following models~\cite{touvron_llama_2023,touvron_llama2_2023,ye_chatgpt_2023}, and the InstructGPT-style instruction tuning paradigm~\cite{ouyang_instructgpt_2022}, mechanisms for expressing and managing explicit short-term memory have evolved from static configuration to dynamic interaction, becoming increasingly structured and flexible.
However, explicit short-term memory in LLMs is physically constrained by context window length. When handling lengthy texts or multi-turn dialogues, models often encounter truncation of early content and memory fading, leading to diminished semantic coherence or loss of key information~\cite{liu_lost_2024,dong_longsurvey_2024}. Recent research has attempted to alleviate these bottlenecks through longer windows, external retrieval, or more efficient caching, yet the capacity of explicit short-term memory remains a key limiting factor in real-time comprehension and interaction.

\paragraph{\textbf{Explicit Long-term Memory in LLMs}}
Unlike short-term memory dependent on context windows, LLMs' explicit long-term memory emphasizes sustained access to external non-parametric knowledge, with a focus on optimizing memory organization structures and retrieval strategies.
Early research focused on identifying effective retrieval mechanisms for recalling relevant content from standalone external memory stores. Common approaches include off-the-shelf retrievers such as BM25~\cite{DBLP:journals/ftir/RobertsonZ09}, Dense Passage Retrieval (DPR)~\cite{DBLP:conf/emnlp/ReimersG19}, and hybrid retrieval methods~\cite{Ensemble-Retriever}.
However, such retrieve-then-generate approaches impose an inherent bottleneck in integrating retrieved content into model reasoning. Thus, some studies have explored tighter coupling of retrieval with inference. Non-parametric language models (NPLMs) such as kNN-LMs~\cite{khandelwal_knnlm_2019,pozzobon_goodtriever_2023} propose a linear fusion of neural language models (e.g., Transformers) with k-nearest-neighbor retrieval. At each prediction step, they retrieve top-matching context chunks from memory and blend their influence into the model’s output distribution to improve reference fidelity.

Due to the limited representational capacity of flat memory structures, optimizing retrieval alone often fails to surpass performance ceilings. As a result, research has increasingly shifted toward enhancing memory organization itself. Traditional key-value formats have gradually evolved into more hierarchical and relational structures, such as tree-based~\cite{DBLP:journals/corr/abs-2310-05029} and graph-based formats~\cite{DBLP:journals/corr/abs-2404-16130,DBLP:journals/corr/abs-2410-05779}.
To further represent diverse memory relationships, researchers have introduced heterogeneous graphs~\cite{xu2025noderagstructuringgraphbasedrag,yang2025heteragheterogeneousretrievalaugmentedgeneration} and hypergraph structures~\cite{DBLP:journals/corr/abs-2503-21322}, enabling unified modeling and dynamic control of varied knowledge types and complex semantic links. These advances greatly enhance the expressive power and generalization of memory networks. To endow LLMs with structured, dynamic, and persistent memory, Zep~\cite{DBLP:journals/corr/abs-2501-13956} builds on GraphRAG~\cite{DBLP:journals/corr/abs-2404-16130} by adding timeline modeling to track memory evolution over time. A-MEM~\cite{DBLP:journals/corr/abs-2502-12110} draws from dynamic memory networks to support automatic memory linking and semantic updating, allowing LLM memory to evolve across multi-turn interactions.

\subsection{Stage 2: Development of Human-like Memory}

To enhance the memory capabilities of LLMs in complex tasks, some studies have drawn inspiration from human memory mechanisms and knowledge management methods, proposing various forms of human-like memory.

In the early stages of human-like memory research, the focus was on simulating the structural and functional mechanisms of human memory.
One representative early work is the HippoRAG series of models\cite{DBLP:conf/nips/GutierrezS0Y024,DBLP:journals/corr/abs-2502-14802}, inspired by the "hippocampal indexing theory" in human long-term memory.
The model integrates LLMs, knowledge graphs, and the Personalized PageRank algorithm to emulate the roles of the neocortex and hippocampus in memory, achieving more efficient knowledge integration and retrieval.
Memory$^3$\cite{memory3_Yang_2024}, inspired by the hierarchical structure of human memory, makes the KV-cache in the attention mechanism explicit as a memory carrier for the model.
This approach offers a lower-cost alternative to parameter storage or traditional RAG, significantly reducing the resource consumption for training and inference.

As research advanced, system designs began emphasizing human-like behavior and function, simulating how humans actually use memory.
For instance, PGRAG\cite{DBLP:journals/corr/abs-2405-16933} mimics the act of note-taking during reading, automatically generating mind maps as explicit long-term memory to enhance organization and durability.
Second-Me\cite{wei2025ainativememory20second} proposes a multi-level architecture centered on human-like memory behaviors, emphasizing experience-driven personalized retrieval.
The system consists of three layers: L0 retains raw data for completeness; L1 enhances organization and retrievability through structured natural language; L2 internalizes user preferences via parameter tuning, enabling associative reasoning similar to humans.
AutoGen\cite{DBLP:journals/corr/abs-2308-08155} introduces a multi-agent framework to simulate human group collaboration, forming a dialog ecosystem of interacting agents.
Each agent has distinct roles, and they collaborate through dialog to share information and accomplish complex tasks like mathematical reasoning, information retrieval, and code generation.

\subsection{Stage 3: Tool-based Memory Management}

With the growing understanding of memory in LLMs, researchers have begun exploring explicit manipulation of knowledge, pushing memory management from implicit representations toward tool-based interfaces.

This stage witnessed the emergence of standardized frameworks for memory editing, enabling users to dynamically update the model’s semantic behavior through insert, modify, and delete operations. For example, early approaches like EasyEdit\cite{zhang_easyedit_2024,xu_easyedit2_2025} offer unified interfaces to manipulate model parameters and hidden states for fine-grained control.
Another representative line of work is Mem0\cite{chhikara_mem0_2025}, which targets the context window bottleneck by introducing external memory modules maintained through extract-update workflows. Follow-ups to Mem0 even structure conversational memory into graphs to enable richer semantic modeling and long-term evolution.
Among these, Letta\cite{packer_memgpt_2024} stands out as a system-oriented attempt. It draws inspiration from traditional operating systems by modularizing context and introducing function-style paging for dynamic memory access.

However, most work in this stage remains limited to interface-level utilities. While tool-based management introduces basic CRUD operations, it lacks systematic modeling and governance of memory as a core resource—making it insufficient for tasks requiring memory evolution, coordination, or security.

\subsection{Stage 4: Systematic Memory Governance}
Although tool-based management introduces explicit memory operation interfaces, it essentially patches implicit mechanisms.
CRUD capabilities alleviate short-term issues but fall short of addressing systemic challenges like memory evolution, access control, and version management.
Just as system calls alone cannot build a complete OS, "tooling" memory lacks a sustainable and scalable governance architecture.

To overcome the limitations of tool-based management, we propose \textbf{\textsc{MemOS}}, a memory operating system purpose-built for LLMs, marking the entry into the stage of systematic memory governance.
\textsc{MemOS} treats memory units as first-class resources and builds upon operating system design principles to introduce comprehensive governance mechanisms including scheduling, layering, API abstraction, permission control, and exception handling.
Unlike the tool-based phase, \textsc{MemOS} not only enables operations but also emphasizes the evolution and integration of memory across tasks, sessions, and agent roles.
With core modules such as MemScheduler, Memory Layering, and Memory Governance, \textsc{MemOS} enables unified scheduling and behavior-driven evolution of heterogeneous memory types—building a long-term cognitive structure essential for AGI.
We envision the “memory-as-OS” paradigm pioneered by \textsc{MemOS} as the infrastructural backbone for future general-purpose agents, enabling sustainable knowledge accumulation and self-evolution.

\section{\textsc{MemOS} Design Philosophy}
\subsection{Vision of \textsc{MemOS}}

As AGI advances toward increasingly complex systems involving multiple tasks, roles, and modalities, LLMs must go beyond merely ``understanding the world''—they must also ``accumulate experience,'' ``retain memory,'' and ``continuously evolve.'' However, current mainstream LLM architectures lack systematic support for memory as a core intelligence capability: knowledge is rigidly encoded in parameters, context cannot be preserved across sessions, personalization cannot be retained, and knowledge updates are prohibitively expensive. We argue that the next-generation LLM architecture must adopt a memory-centric design paradigm.

\begin{wrapfigure}{r}{0.5\textwidth}
    \centering
    \vspace{-1.5em}
    \includegraphics[width=0.48\textwidth]{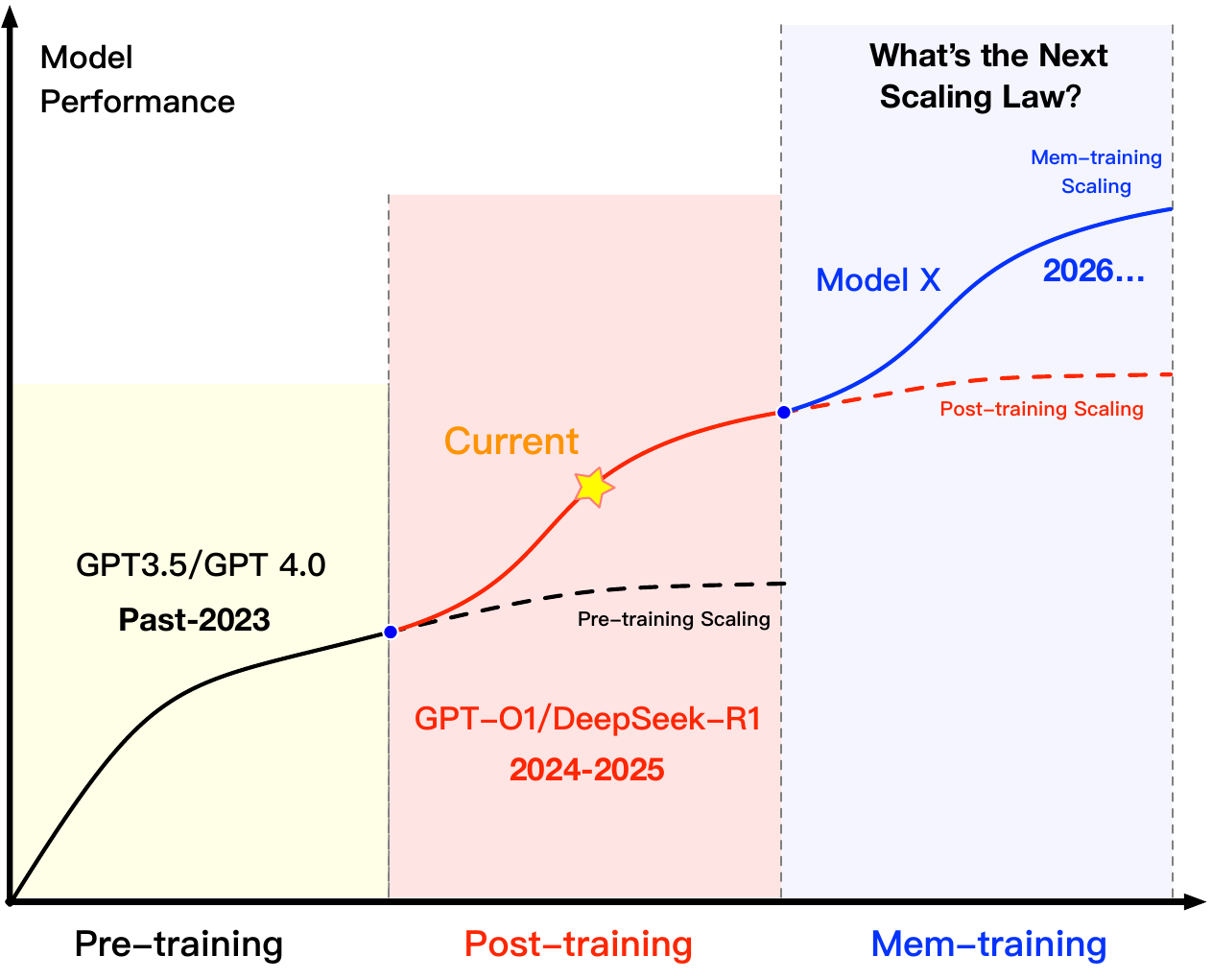}
    \caption{Phased transitions in model performance: from pretraining and post-training to the Mem-training stage. \textsc{MemOS} serves as the foundational infrastructure enabling the next era of scaling laws.}
    \label{fig:memos_scaling}
    \vspace{-1em}
\end{wrapfigure}

As shown in Figure~\ref{fig:memos_scaling}, model performance is approaching the upper limits predicted by traditional scaling laws. The prevailing research paradigm is transitioning from data- and parameter-centric pretraining to post-training, which emphasizes reinforcement alignment and instruction tuning \cite{zhou_lima_arxiv23}. Yet this shift faces two major challenges: diminishing returns and growing system complexity. To unlock the next leap in capability, we must transcend the current paradigm by incorporating continuous memory modeling and dynamic memory scheduling—thereby enabling long-term knowledge accumulation, task adaptation, and behavioural evolution.

Beyond the temporal benefits of continual learning, memory training also introduces a spatial scaling effect. Thousands of heterogeneously deployed model instances can gather experience in situ and exchange compact memory units—rather than expensive parameters or gradients—to build a collective knowledge base. This memory-parallel regime blurs the line between training and deployment, effectively extending data parallelism to a society-scale, distributed intelligence ecosystem.
Two technical challenges arise:  
(1) efficient knowledge exchange across highly heterogeneous environments, and  
(2) strict governance that protects private or sensitive data while maximising shared utility.  

We therefore advocate a memory-centric training strategy—the Mem-training Paradigm. Instead of relying solely on sporadic parameter updates, Mem-training drives continuous evolution through explicit, controllable memory units. Unlike traditional workflows that modify the model only during pretraining or fine-tuning, Mem-training allows knowledge to be collected, re-structured, and propagated at runtime, enabling self-adaptation across tasks, time horizons, and deployment environments.

In this paradigm, "training" is no longer limited to large-scale corpora but extends to dynamic knowledge accumulation via continuous interaction with users and the environment. The focus shifts from how much knowledge the model learns once to whether it can transform experience into structured memory and repeatedly retrieve and reconstruct it. \textsc{MemOS} serves as the system-level foundation for this paradigm, enabling end-to-end capabilities in memory generation, scheduling, fusion, and updating.

Our vision is for \textsc{MemOS} to become the foundational memory infrastructure for next-generation intelligent agents, with its core mission expressed through the following three pillars:

\begin{itemize}
    \item \textbf{Memory as a System Resource}: Abstract memory from a latent, internal dependency into a first-class, schedulable, and manageable resource. Build memory pathways that span agents, users, applications, and sessions, breaking down "memory silos" across platforms, significantly reducing memory management complexity, and improving the effectiveness and efficiency of memory access.

    \item \textbf{Evolution as a Core Capability}: Enable continuous learning, structural reorganization, and task transfer throughout long-term memory usage. Build a co-evolutionary infrastructure for models and memory, allowing LLMs to self-adapt and upgrade in response to changing tasks, environments, and feedback—achieving truly sustainable, evolving intelligence.

    \item \textbf{Governance as the Foundation for Safety}: Provide lifecycle-wide memory governance mechanisms including access control, versioning, provenance auditing, and more. Ensure controllability, traceability, and explainability of memory, laying the groundwork for secure, trustworthy, and compliant intelligent agent systems.
\end{itemize}

We believe that just as traditional operating systems laid the foundation for modern computing by unifying computation and storage management, \textsc{MemOS} will elevate memory to a core system resource, forming an indispensable foundation for both general-purpose and embodied intelligent agents. This will drive a paradigm shift from reactive, perception-based systems to memory-driven, evolving agents.

\subsection{From Computer OS to Memory OS}

In traditional computing systems, the operating system (OS) centrally manages key hardware resources—such as the central processing unit (CPU), memory, storage devices, and peripherals—to support efficient execution and stable operation of applications. The OS’s abstraction of resources, unified scheduling, and lifecycle governance serve as the foundation for the scalability and reliability of modern computing infrastructures.

As large language models (LLMs) scale in inference and application complexity, both internal and external memory resources—ranging from static parameter memory to runtime activation memory and dynamically retrieved explicit memory modules—exhibit increasingly dynamic and heterogeneous behavior. These memory forms are not only foundational to inference but also continuously evolve with task shifts and knowledge updates. Therefore, LLMs similarly require a systematic resource management framework akin to traditional operating systems, enabling standardized abstraction, dynamic scheduling, and autonomous lifecycle governance of memory.

\textsc{MemOS} proposes a design philosophy for the unified and systematic management of memory resources in LLMs, drawing extensively on mature mechanisms from traditional OS domains such as resource scheduling, interface abstraction, access control, and fault handling. Table~\ref{table:mapping} illustrates the mapping between classical OS components and \textsc{MemOS} modules: \textsc{MemOS} coordinates inference and memory block scheduling via the LLM Core and MemScheduler, manages hierarchical memory through Memory Layering and MemStore, offers standardized API abstraction through MemAPI and Backend Adapter, enforces security and access governance through Memory Governance, and supports monitoring and anomaly detection through the Memory Observability framework. These modules work in concert to adapt traditional resource management principles to the evolving demands of memory in LLMs.

\input{tab/osmapping}

\section{Memory Modeling in \textsc{MemOS}}

\subsection{Types of Memory in \textsc{MemOS}}

The concept of hierarchical memory was originally introduced in our prior work Memory\textsuperscript{3}\cite{memory3_Yang_2024}, which proposed a distinction between explicit and implicit memory paths in LLMs and investigated their interaction mechanisms.

Building on this foundation, \textsc{MemOS} systematizes the idea by delineating three core memory types—\textbf{Plaintext Memory}, \textbf{Activation Memory}, and \textbf{Parameter Memory}—that together reflect a full semantic evolution trajectory from perception to consolidation.

To coordinate scheduling and evolution across heterogeneous memory types, \textsc{MemOS} introduces the \hyperlink{memcube}{\textbf{MemCube}}—a unified abstraction that standardizes memory representation, lifecycle management, cross-modal fusion, and dynamic memory state transitions.
Its design is inspired by the controllable externalization proposed in Memory\textsuperscript{3}, while advancing it into a composable and schedulable memory substrate suitable for intelligent agent construction.
This design forms the semantic memory backbone of \textsc{MemOS}, enabling seamless integration and transformation of multiple memory types during inference.

\paragraph{\textbf{Plaintext Memory}} 
Plaintext memory refers to explicit, dynamically retrieved knowledge modules accessed via external interfaces—editable, traceable, and storable independently.
Examples include retrieved passages, structured graphs, and prompt templates.
Injected into model input, it bypasses the limitations of parameter capacity and context window size.
It enables rapid knowledge updates, task customization, and user personalization.

\textsc{MemOS} encapsulates plaintext memory into tunable \texttt{MemCube}s, with lifecycle control, access policies, and version tracking.
It supports graph-structured and multimodal memory, contextual fingerprinting, and timestamp-based loading.
Plaintext memory is not merely an external plugin. \textsc{MemOS} deeply integrates it into the inference loop, enabling interaction with activation memory.
High-frequency plaintext can be transformed into activation paths, achieving dynamic externalization and internalization of knowledge.
To enhance scheduling efficiency and long-term evolvability, \textsc{MemOS} manages plaintext memory in a hierarchical graph structure organized by task–concept–fact paths.
Task parsing combined with semantic similarity and topic-aware strategies enables structured query routing and prioritized retrieval.
It supports conflict detection, deduplication, versioning, and forgetting policies to maintain memory quality and evolution.

Plaintext memory is particularly suited for fact-heavy, personalized, and multi-agent tasks—serving as a core enabler of transparent and collaborative intelligence.

\paragraph{\textbf{Activation Memory}}
Activation memory consists of intermediate states generated during inference, with the KV-cache as the central structure.
It retains key-value representations of context, enabling efficient long-range dependency modeling and recursive reasoning.
It supports instant contextual response and reusable inference pathways through cache-stable behaviors.
Other elements include hidden states ($h^l_i$) and attention weights ($\alpha^l_{ij}$), comprising the model’s runtime semantic perception.
These are characterized as short-term, dynamic, and implicitly activated.

\textsc{MemOS} offers unified scheduling and lifecycle management for activation memory.
It enables lazy loading, selective freezing, and priority-driven adjustments.
Frequent KV patterns are cached to form low-latency “instant memory paths”.
Beyond KV patterns, strategic behaviors that are repeatedly triggered can also be abstracted into persistent memory structures, such as steering vectors or semantic templates.
KV memory proves valuable in multi-turn dialogue, code assistance, and runtime safety management. 
For instance, in medical agent systems, stable and frequently accessed knowledge—such as patient histories, routine diagnostic procedures, or clinical commonsense—can be abstracted into cached KV segments, enabling rapid recall and minimizing redundant decoding.
It is essential for maintaining contextual continuity, stylistic coherence, and precise response control.

\paragraph{\textbf{Parameter Memory}}
Parameter memory refers to knowledge and capabilities encoded in the model’s fixed weights. It serves as the primary repository of long-term semantic knowledge within the model.
It encodes deep representations of linguistic structure, commonsense knowledge, and general semantics—typically instantiated as feedforward weight matrices (e.g., $W^l_{\text{MLP}}$) and attention key/value matrices (e.g., $W^l_K$, $W^l_V$).
Unlike other memory types, parameter memory is activated implicitly without retrieval or explicit context, forming the foundation for zero-shot inference, general QA, and language generation.

In \textsc{MemOS}, parameter memory includes both pre-trained linguistic and world knowledge and can be modularly enhanced via lightweight fine-tuning methods such as LoRA or adapters.
\textsc{MemOS} enables distilling domain-specific knowledge into parameter blocks, loadable as ``capability modules" (e.g., summarization expert, legal assistant, style generator).
While offering strong expressivity and high efficiency, parameter memory suffers from high update costs, limited customizability, and poor interpretability.
To address this, \textsc{MemOS} links parameter memory with plaintext and activation memories.
For instance, frequently used and structurally stable plaintext may be distilled into parametric form for embedded efficiency.
Conversely, outdated or inconsistent parameter memory can be backpatched by reverting to plaintext.
Parameter memory is ideal for capability-centric agents, such as legal advisors, financial auditors, technical writers, or summarizers, or as composable “capability plugins”.
Compared with frequently updated plaintext or transient activation memory, it better supports long-term, structurally stable capabilities.

\begin{figure}[htp]
    \centering
    \includegraphics[width=0.8\linewidth]{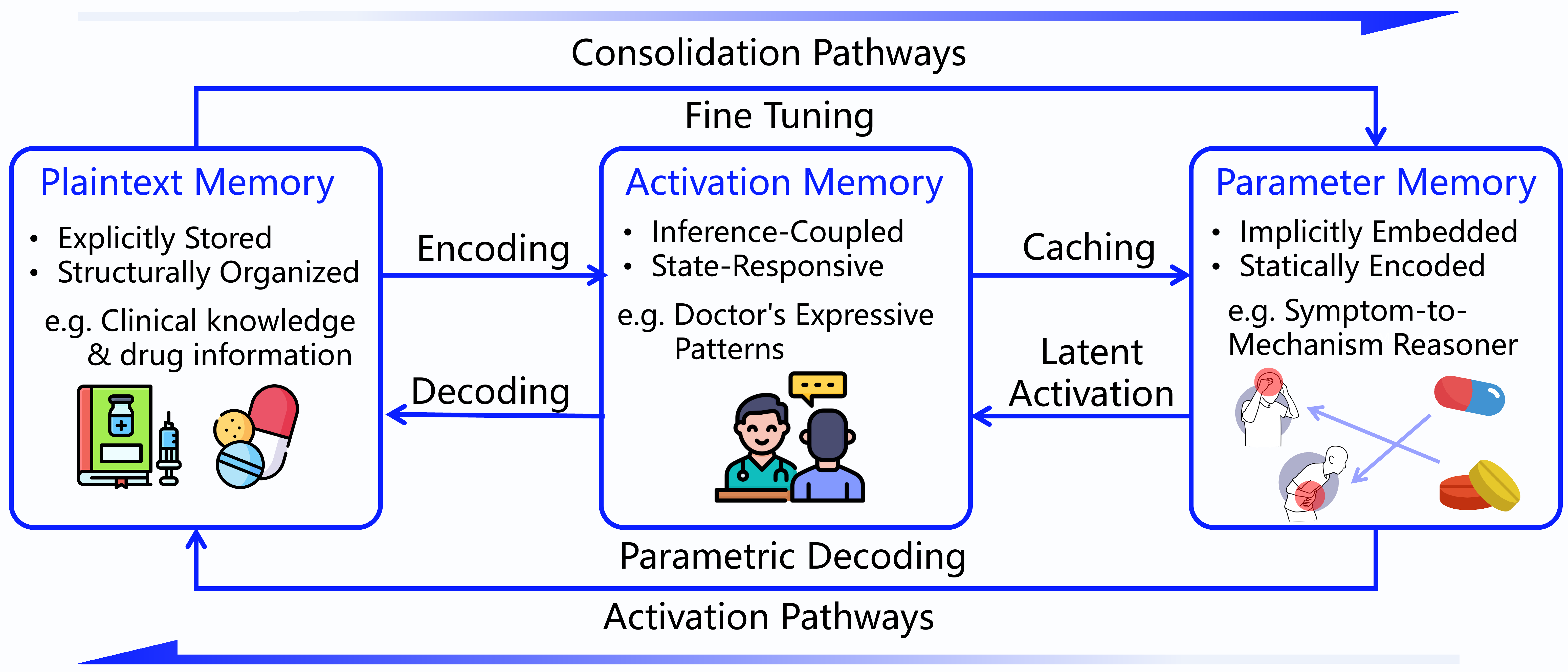}
    \caption{Transformation paths among three types of memory, forming a unified, controllable, and evolvable memory space.}
    \label{fig:memories}
\end{figure}

\subsection{Memory Cube (\texttt{MemCube}) as a Core Resource}

\begin{figure}[htp]
    \centering
    \includegraphics[width=1.0\linewidth]{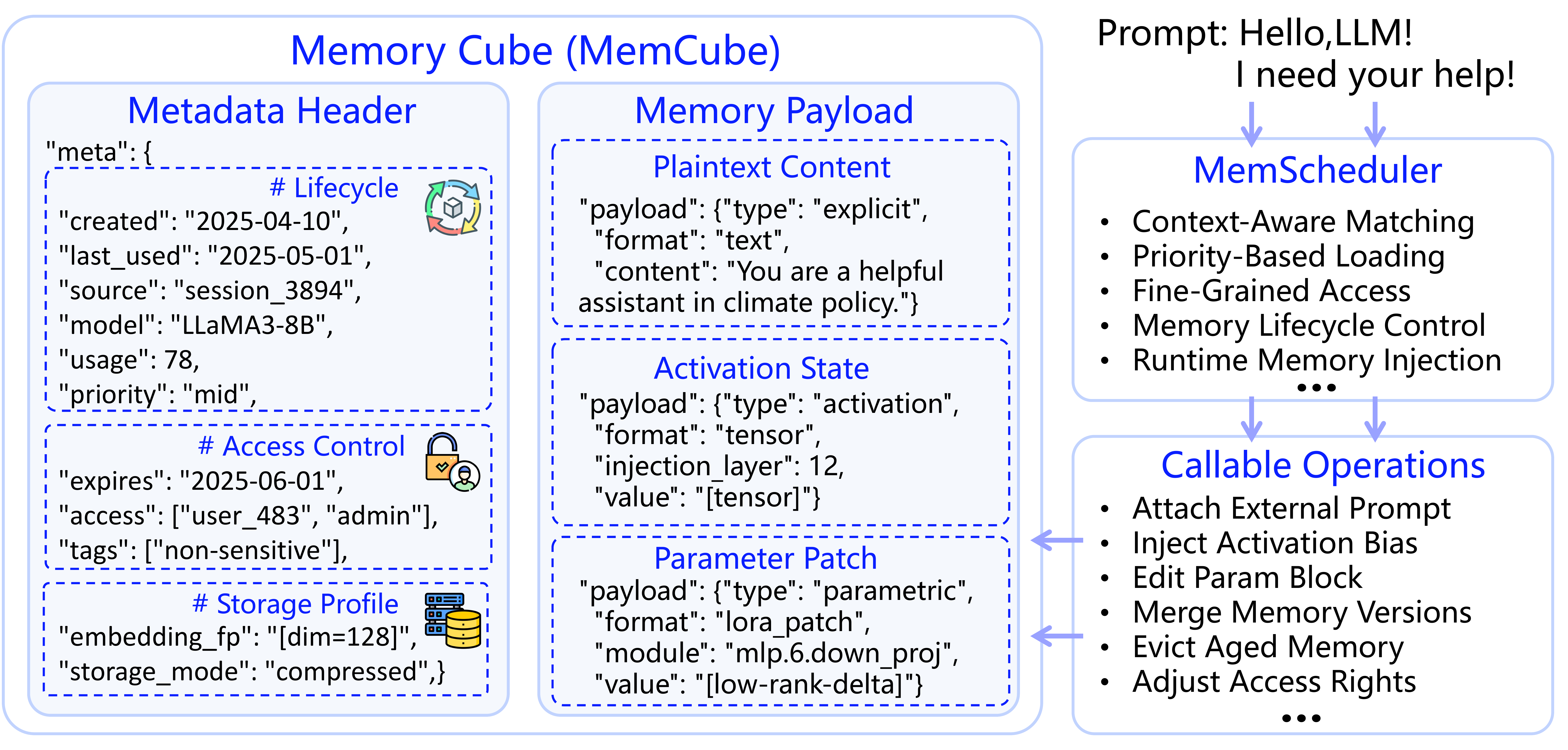}
    \caption{\texttt{MemCube}: A unified encapsulation structure for heterogeneous memory scheduling. Each \texttt{MemCube} consists of a structured Metadata Header (supporting lifecycle, permission, and storage policy) and a Memory Payload (encapsulating plaintext, activation states, or parameter deltas). It is the minimal memory unit within \textsc{MemOS} that can be scheduled and composed for downstream reasoning.}
    \label{fig:MemCube}
\end{figure}

In \textsc{MemOS}, the foundation of a unified and structured memory management system lies in the standardized abstraction and system-level governance of heterogeneous memory resources. To this end, we propose the \hypertarget{memcube}{\textbf{\texttt{Memory Cube (MemCube)}}} as a universal encapsulation unit for memory resources (see Figure~\ref{fig:MemCube}).

Memory in LLMs is highly diverse, spanning long-term knowledge embedded in model parameters, intermediate activation states generated during inference, and externally injected structured knowledge fragments (e.g., retrieved passages, knowledge graph nodes). These resources differ significantly in origin, lifecycle, representation, and scheduling method, making unified control, evolution, and governance a systemic challenge.

The design of \texttt{MemCube} aims to encapsulate all memory types as unified scheduling units, each with standard interfaces, behavioral properties, and governance strategies. Each \texttt{MemCube} instance consists of two components: the \texttt{Memory Payload}, which contains the semantic content, and the \texttt{Metadata}, which encodes identity, control, and behavioral metrics. These metadata elements serve as foundational interfaces for \textsc{MemOS} scheduling and governance and as central anchors for long-term system evolution, task adaptation, and security control.

The metadata of each \texttt{MemCube} is categorized into three groups: \textbf{descriptive identifiers}, \textbf{governance attributes}, and \textbf{behavioral usage indicators}. Together, these enable full-spectrum memory management across structural identification, access control, and behavioral evolution. We elaborate below on their motivations, components, and system-level implications.

\paragraph{\textbf{Descriptive Identifiers}} define each memory block's identity, classification, and organization. Unified memory scheduling at scale relies on precise identification of these “semantic fingerprints.” \texttt{MemCube} embeds key fields such as: \textbf{Timestamp}, indicating creation or last update for lifecycle modeling; \textbf{Origin Signature}, identifying whether the memory comes from inference extraction, user input, external retrieval, or parameter finetuning; and \textbf{Semantic Type}, specifying its use (e.g., task prompt, fact, user preference) to support semantic composition. These jointly enable layered memory structuring and contextual navigation.

\paragraph{\textbf{Governance Attributes}} provide systemic controls for memory access, security, and scheduling. In dynamic, multi-user, long-running systems, default model reasoning is insufficient for robust memory governance. \textsc{MemOS} defines a comprehensive rule set per memory unit, including: \textbf{Access Control} (read/write/share scope), \textbf{Lifespan Policy} (TTL or decay rules), \textbf{Priority Level} (for scheduling), and \textbf{Compliance \& Traceability} (e.g., sensitivity tags, watermarks, logs). Together, they form the memory governance kernel—critical for system stability, transparency, and accountability.

\paragraph{\textbf{Behavioral Usage Indicators}} reflect real-time memory usage during inference, enabling “value-driven” scheduling and cross-type transformation. Unlike static labels, these runtime metrics empower adaptive orchestration of memory.

\textbf{Access Patterns}, such as frequency and recency, inform whether a memory is “hot” or “cold” during inference. \textsc{MemOS} uses this to adjust caching priority—for example, promoting high-frequency plaintext memory into fast-access layers to reduce latency.

These indicators also support \textbf{Cross-Modality Memory Transformation}, allowing dynamic transitions across memory types:
\begin{itemize}
    \item \textbf{Plaintext $\Rightarrow$ Activation:} Frequently used plaintext memory can be pre-transformed into activation vectors or attention templates for faster decoding.
    
    \item \textbf{Plaintext/Activation $\Rightarrow$ Parameter:} Stable knowledge across tasks can be distilled into parameter modules, internalized as efficient capability plugins.

    \item \textbf{Parameter $\Rightarrow$ Plaintext:} Cold or outdated parameters can be offloaded into external plaintext storage to increase flexibility and reduce structural overhead.
\end{itemize}

To support such transformations, \textsc{MemOS} introduces \textbf{Policy-Aware Scheduling}: the system dynamically adjusts a memory block’s tier and format based on usage frequency, contextual dependency, and task fit—enabling layered memory evolution.
Additionally, each memory is associated with a \textbf{Contextual Fingerprint}, a lightweight semantic signature for fast retrieval and task alignment.
A \textbf{Version Chain} logs each memory’s modification history and derivation lineage, enabling version control, conflict resolution, and rollback.
These behavioral metrics allow \textsc{MemOS} to perceive the “value” of memory, forming the basis for adaptive scheduling, memory transformation, and knowledge evolution. As a result, memory becomes a self-regulating and self-evolving intelligent resource unit.
Through the coordinated design of these three metadata types, \texttt{MemCube} enables structured abstraction, permissioned control, and behavior-driven evolution of heterogeneous memory resources.

\section{Architecture of \textsc{MemOS}}
\label{sec:architecture}

\begin{figure}[ht]
    \centering
    \includegraphics[width=1.0\linewidth]{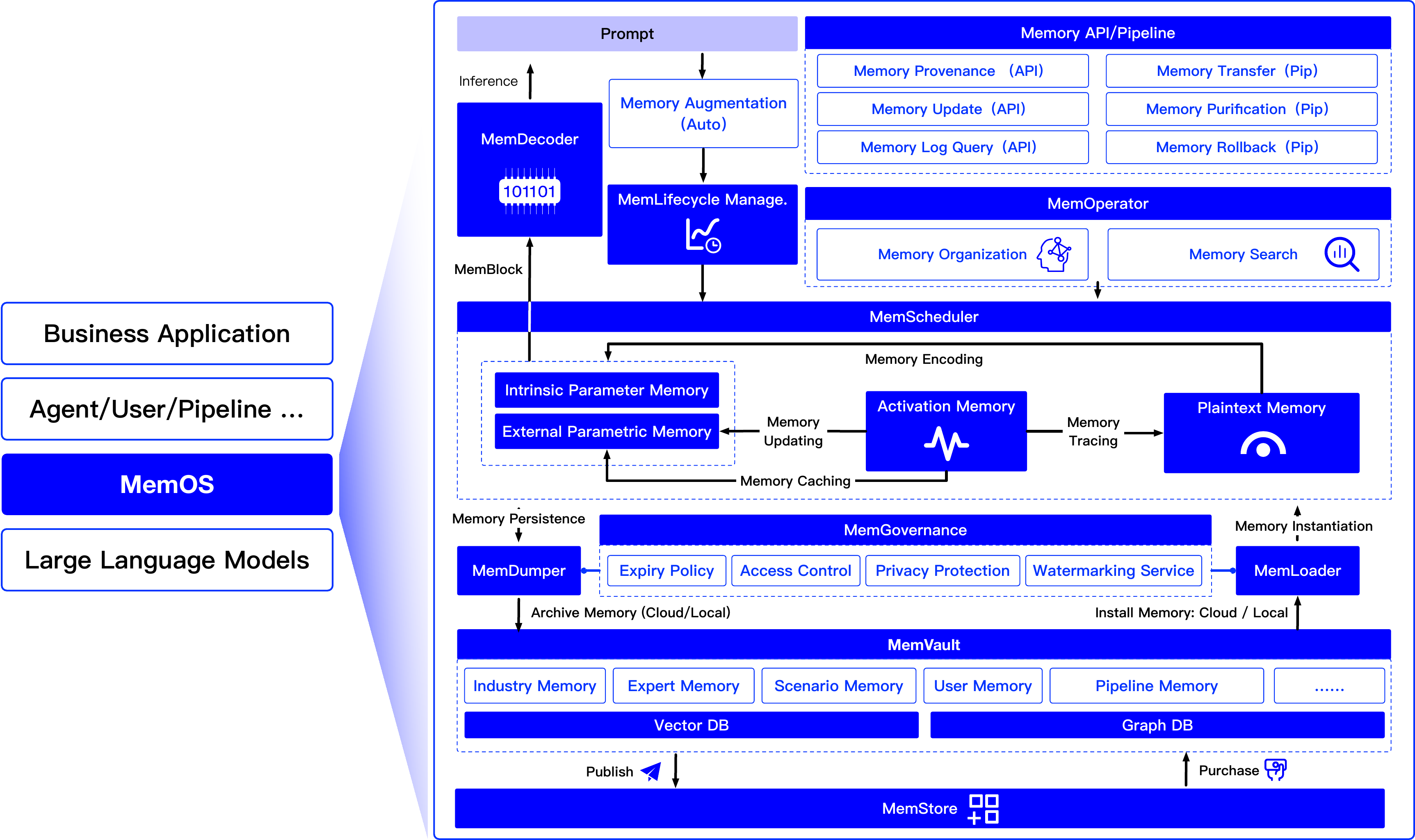}
    \caption{Overview of the \textsc{MemOS} framework. The architecture illustrates the full pipeline from user input through semantic parsing and API abstraction in the interface layer, to memory scheduling and lifecycle control in the operation layer, and finally interaction with the infrastructure layer for memory injection, retrieval, and governance. The unified data structure, \texttt{MemCube}, serves as the foundation for dynamic memory flow throughout model execution.}
    \label{fig:framework}
\end{figure}

\subsection{Overview: Three-layer Architecture of \textsc{MemOS}}
\textsc{MemOS} adopts a modular three-layer architecture to support efficient invocation, dynamic scheduling, and compliant governance of complex memory tasks (see Figure~\ref{fig:framework}). It consists of the Interface Layer, Operation Layer, and Infrastructure Layer, each with distinct responsibilities and collaborative interfaces—together building a unified execution and governance framework for heterogeneous memory types that enables robust intelligent agent performance across complex tasks.

\paragraph{\textbf{Memory Interface Layer}}
The interface layer interacts with users or upstream systems and serves as the entry point for all memory operations.
It provides a standardized \texttt{Memory API} suite that supports querying, writing, updating, transferring, and composing memory units.
All user requests are parsed by the interface layer into specific memory manipulation commands.
The built-in \texttt{MemReader} module plays a central role in this process.
It converts natural language inputs into structured memory operation chains, extracting time expressions, task intents, contextual anchors, and memory scopes.
For instance, given a request like “Summarize my meeting notes from last month,” \texttt{MemReader} extracts the time range (last month), memory type (meeting notes), and output target (summary), and formulates a labeled \texttt{MemoryQuery} with proper window parameters.
In multi-turn conversations, \texttt{MemReader} uses context to infer omitted details, ensuring consistency in memory invocation.
This layer also performs permission checks, parameter encapsulation, and call sequence management. It coordinates with \texttt{MemGovernance} to validate the compliance and traceability of every operation.

\paragraph{\textbf{Memory Operation Layer}}
The operation layer serves as the control center of \textsc{MemOS}, organizing, planning, and scheduling memory resources during inference.
Its core components include \texttt{MemOperator}, which builds tag systems, semantic indexes, and graph-based topologies across heterogeneous memory types and contexts, facilitating efficient retrieval and contextual adaptation.
\texttt{MemScheduler} selects appropriate memory types (e.g., Plaintext, activation, parameter ) based on task intent and context, and dynamically plans invocation order and integration strategy to optimize for low latency and task relevance.
\texttt{MemLifecycle} tracks the lifecycle transitions of each memory unit—creation, activation, expiration, and reclamation—to ensure memory resource controllability and freshness.
In a multi-turn QA or complex dialogue, the operation layer first retrieves relevant memory (e.g., user preferences, past conversations, external structured documents) via \texttt{MemOperator}, determines the optimal invocation path via \texttt{MemScheduler}, and updates memory states using \texttt{MemLifecycle}.
Thanks to this design, memory becomes a dynamic, context-aware resource rather than a static data fragment.

\paragraph{\textbf{Memory Infrastructure Layer}}
The infrastructure layer handles storage, security, migration, and flow of memory data, serving as the foundation for reliable system execution.
\texttt{MemGovernance} enforces access control, retention policies, audit logging, and sensitive content handling.
\texttt{MemVault} manages multiple memory repositories (e.g., user-specific, domain knowledge, shared pipelines) and provides standardized access interfaces.
\texttt{MemLoader} and \texttt{MemDumper} enable memory import/export and cross-platform synchronization.
\texttt{MemStore} provides a publish-subscribe mechanism for open memory sharing among multiple agents.
In organizational QA systems, for instance, a locally updated memory entry can be validated and synchronized to a central memory hub, becoming available to authorized users.

Together, these three layers form the complete memory operation loop in \textsc{MemOS}—from task input to execution scheduling to governance and archival. The standard interface decoupling allows rapid iteration and extensibility, laying the foundation for multi-model, multi-task, and cross-platform memory sharing in future intelligent systems.

\subsection{Execution Path and Interaction Flow of \textsc{MemOS}}

\begin{figure}[htp]
    \centering
    \includegraphics[width=1.0\linewidth]{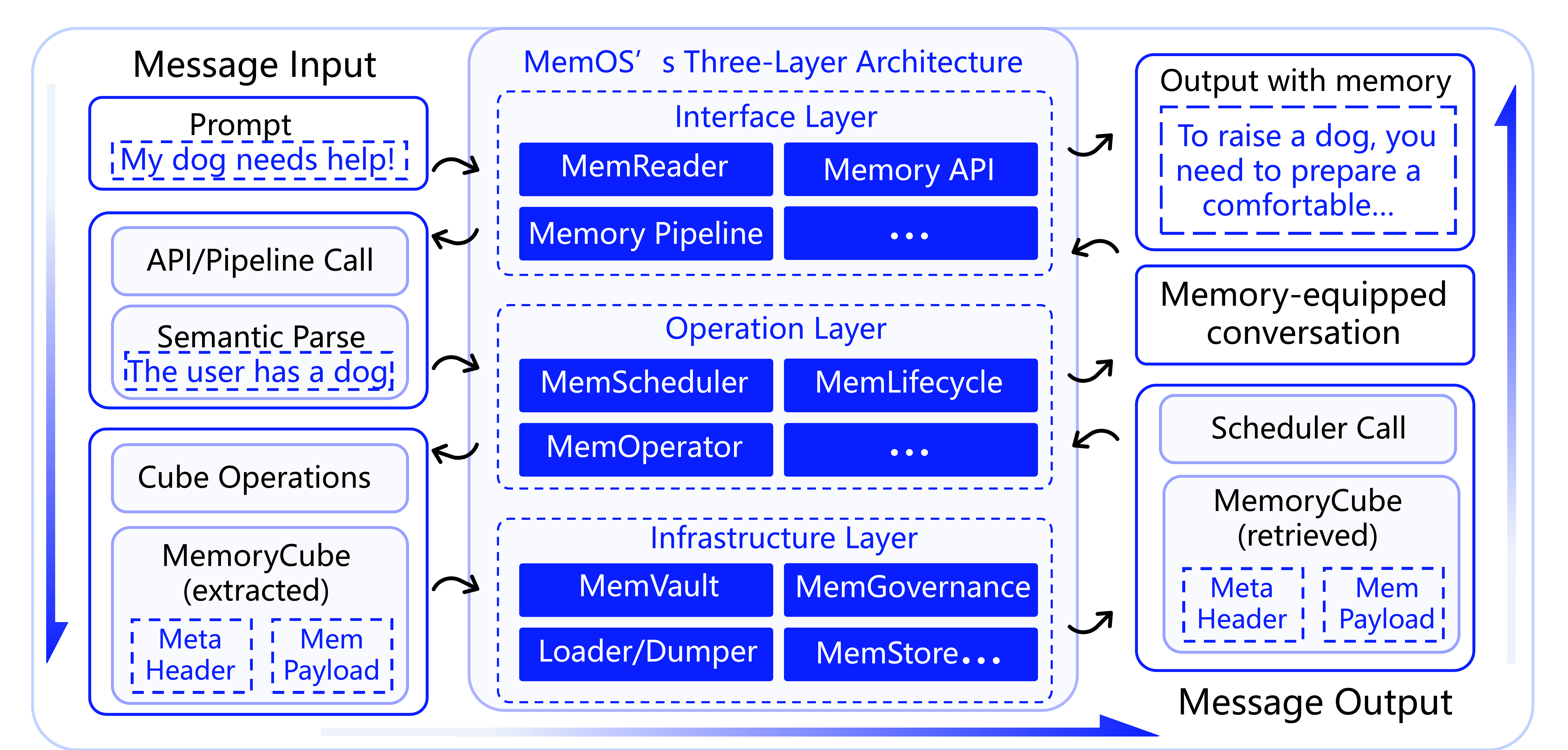}
    \caption{Overview of MemOS architecture and memory interaction flow. The system is composed of the interface layer, operation layer, and infrastructure layer. From left to right, it shows the complete memory processing pipeline from user input to parsing, scheduling, injection, and response generation. Each stage corresponds to coordinated module invocation, with MemoryCube serving as the carrier across layers for structured, governable, and traceable memory lifecycle management.}
    \label{fig:overview}
\end{figure}

The execution of \textsc{MemOS} is triggered by either user interaction or automated tasks. It follows a closed-loop process through input parsing, memory scheduling, state management, and storage archiving(Figure~\ref{fig:overview}). 

\paragraph{\textbf{Prompt Input and Memory API Packaging}}
System execution begins with a user-issued natural language prompt or an automatically triggered task. The interface layer processes the input through the built-in \texttt{MemReader} module, which identifies task intent, time scope, topic entities, and contextual anchors to determine if memory access is involved. If so, \texttt{MemReader} converts the prompt into a structured \texttt{MemoryCall}, including the caller ID, context scope, memory type, access intent, and time window. This is encapsulated into a standardized \texttt{Memory API} request and passed to the operation layer for execution.
For example, in a healthcare scenario, when a patient inputs, “Please retrieve my inpatient records from last year,” \texttt{MemReader} identifies the time range (last year), topic tag (diagnostic records), contextual anchor (hospitalization period), and intent (historical query), and generates a structured \texttt{MemoryCall}, which proceeds to the memory retrieval and scheduling pipeline.

\paragraph{\textbf{Memory Retrieval and Organization}}
The \texttt{MemOperator} in the operation layer uses intent and context info from the \texttt{Memory API} to perform semantic matching and organize memory units. It constructs task-specific indexes (user preferences, anchors, keyword vectors) and memory graphs (temporal chains, entity relations, dependencies) to filter relevant candidates.
For instance, if a patient asks the system to reference past cases for diagnosis, the operator retrieves memory blocks with symptom keywords, treatment periods, and associated physician notes to construct a structured retrieval path.

\paragraph{\textbf{Memory Scheduling and Activation}}
After the candidate set is identified, \texttt{MemScheduler} optimizes memory selection using metrics like contextual similarity, access frequency, temporal decay, and priority tags. It dynamically computes the optimal injection strategy.
In a follow-up appointment, the system injects recent consultation summaries (activation memory), diagnosis templates (parameter memory), and lifestyle advice (plaintext memory), ensuring integrated, semantically coherent support.

\paragraph{\textbf{Lifecycle Modeling and State Transitions}}
Scheduled memory units are passed to \texttt{MemLifecycle} for state management. Each memory item transitions through five states—Generated, Activated, Merged, Archived, and Expired—based on access patterns, time decay, and task labels.
For example, in medical use, generated medication advice starts in "Generated" state. If frequently accessed, it becomes "Activated"; after repeated user confirmations, it is "Merged" into frequent-use suggestions; and eventually archived or expired if unused.

\paragraph{\textbf{Storage Archiving and Access Governance}} 
Evolved memories are archived in \texttt{MemVault} and organized by user, task, or context. Archiving may be triggered by policy, user command, or scheduling, keeping frequently accessed data active and less-used data cold or long-term stored.
The archiving phase also invokes \texttt{MemGovernance} for permission encapsulation and compliance checks. Each memory unit is assigned a set of access control strategies—such as Access Control List (ACL), Time-To-Live (TTL), and conditional activation policies—that determine its availability based on user roles and task context.
For example, a treatment summary may be fully visible to the care team but partially visible to the patient. After redaction and watermarking, it can be registered in \texttt{MemStore} for sharing across institutions.

The full governance and archiving pipeline ensures that all memory units—across diverse modalities and agents—are handled in a structured, transparent, and traceable manner, maintaining compliance and efficiency across collaborative healthcare environments.

\subsection{Interface Layer}

\subsubsection{MemReader}
In \textsc{MemOS}, the first step of any memory operation is interpreting natural language inputs from users or system tasks. This responsibility is handled by the \texttt{MemReader}, which serves as the semantic abstraction module for memory-level reasoning. It parses incoming prompts to extract key memory-related features—such as task intent, temporal scope, entity focus, memory type, and contextual anchors—and outputs a structured intermediate representation.
For example, a prompt like "Remind me what the doctor said about my medication during last year's hospitalization" would be parsed by \texttt{MemReader} into a structured memory access plan: task intent (retrieval), time scope (last year), topic (medication guidance), and context anchor (hospitalization period). This plan is passed downstream as a \texttt{MemoryCall} to be processed by the memory operation layer.
\texttt{MemReader} also supports prompt rewriting, coreference resolution, and dialogue memory slot filling across multi-turn interactions. It functions as both an intent recognizer and memory orchestrator, ensuring the system issues precise and traceable calls to the underlying memory infrastructure.

\subsubsection{Memory API}
The interface layer of \textsc{MemOS} is built around a unified and composable \texttt{Memory API}, which bridges upper-level tasks with backend memory operations. All memory-related actions—including creation, updates, retrieval, and auditing—are performed via standardized APIs that ensure extensibility, composability, and governance.
\textbf{\texttt{Provenance API}} enables provenance tracking by embedding metadata into memory objects at creation or modification time. This includes event triggers, contextual state, model identifiers, and external links. Each memory is tagged with a unique provenance ID that persists throughout its lifecycle. Provenance metadata supports explainability, debugging, access control, and memory lineage tracing.
\textbf{\texttt{Update API}} supports mutation operations such as append, merge, or overwrite. It is version-aware, allowing snapshots and label-based differential writes. Typical use cases include task result logging, user correction, and fine-grained memory consolidation. When paired with \texttt{MemLifecycle}, update operations can trigger state transitions and index refreshes.
\textbf{\texttt{LogQuery API}} allows structured access to memory access logs and execution traces. It supports filtering by timestamp, caller identity, memory type, and operation kind. It is essential for debugging, hotspot analysis, auditing, and governance enforcement. For instance, developers can investigate memory usage that led to faulty responses, or validate whether specific memories were invoked.
All \texttt{Memory API} calls use \texttt{MemoryCube} as their parameter carrier and response format. They support transactional safety, structured status reporting, and are governed by \texttt{MemGovernance}, which enforces access control based on users, roles, models, and tasks.

\subsubsection{Memory Pipeline}
To support complex workflows in enterprise and multi-agent settings, \textsc{MemOS} offers a pipeline-style composition mechanism for chaining memory operations. Developers or agent systems can define a sequence of memory actions—e.g., retrieve → augment → update → archive—and execute them as a cohesive pipeline.
Each pipeline step operates on a shared \texttt{MemoryCube} object, which carries input-output state, metadata, and intermediate artifacts. For example, a medical assistant might define a pipeline that (1) retrieves past medication notes via \texttt{LogQuery}, (2) adds doctor’s latest instructions via \texttt{Update}, (3) tags the memory with a new provenance entry, and (4) archives it post-consultation.
Pipelines support transactional consistency, rollback, and fault isolation. They can be defined declaratively through a domain-specific language (DSL), or constructed programmatically. For agent orchestration, \texttt{MemScheduler} interprets dependencies across steps and coordinates scheduling. Pipeline templates can be reused across agents—e.g., for follow-up generation in customer support, or for diagnosis tracking in clinical triage.

By enabling compositional memory flows, \textsc{MemOS} empowers developers to model higher-level cognition patterns, task-specific knowledge shaping, and auditable memory workflows.

\subsection{Operation Layer}
\subsubsection{MemOperator}

In \textsc{MemOS}, efficient memory organization and accurate retrieval are fundamental to enabling intelligent behavior generation, contextual reasoning, and knowledge reuse. The \texttt{MemOperator} module fulfills this role by structuring memory content both logically and semantically. It incorporates tag-based annotation, graph-based linking, and hierarchical abstraction to support multi-perspective memory modeling. Simultaneously, it provides unified interfaces for hybrid retrieval, serving diverse agents across tasks, models, and user contexts.

\paragraph{Multi-perspective Memory Structuring}
\textsc{MemOS} employs three complementary mechanisms for organizing memory. First, a flexible tagging system allows each memory unit to be annotated with metadata such as topic, source, credibility, and sentiment, supporting both user-defined and model-predicted labels. Second, a knowledge-graph structure treats memory as nodes connected via semantic edges, enabling traversable relations across memory items. Third, a semantic layering scheme segments memory into private, shared, and global layers, facilitating memory isolation and coordinated access across tasks and roles.

\paragraph{Hybrid Retrieval and Dynamic Dispatch}
The \texttt{MemOperator} module supports hybrid retrieval mechanisms that combine symbolic and semantic strategies. Structured retrieval applies rule-based filtering over tags, time spans, Boolean conditions, and access control policies. Semantic retrieval uses embedding-based vector representations to identify contextually relevant memory units via similarity search. These two mechanisms can be composed into complex query expressions—such as tag filters combined with semantic ranking—to serve applications like multi-turn dialogue, question answering, or knowledge integration.

\paragraph{Pipeline Coupling and Caching Strategy}
Retrieved memory units are passed downstream as inputs to execution pipelines, tightly coupled with the \texttt{Memory API} and \texttt{MemoryCube} modules. To minimize latency, \textsc{MemOS} implements a local index caching strategy whereby frequently accessed memory is automatically migrated to high-speed intermediate storage. Cache invalidation is managed by heuristics based on usage frequency and contextual drift, with the \texttt{MemScheduler} module overseeing refresh operations in a dynamic, workload-aware manner.

\paragraph{Task-Aligned Memory Routing}
To address the complexity of real-world tasks, \textsc{MemOS} employs a task-aligned routing mechanism that resolves memory navigation paths based on hierarchical semantic goals. User inputs are decomposed into a topic–concept–fact structure, forming a three-layered task schema. The \texttt{MemoryPathResolver} component then formulates a retrieval strategy that answers three key questions: what to search, where to search, and in what order. This structured approach enhances interpretability, scheduling relevance, and alignment between memory selection and task intent.

\subsubsection{MemScheduler}

\texttt{MemScheduler} is the central memory dispatcher of the \textsc{MemOS} operation layer. Its purpose goes beyond simply "retrieving" stored memories; it dynamically transforms and loads them into the runtime context based on task semantics, call frequency, and content stability.  
Relying on the three memory types defined in \texttt{MemCube}—Activation Memory (KV-Cache), Plaintext Memory, and Parameter Memory—\texttt{MemScheduler} supports classification, transformation, and hierarchical dispatch to deliver adaptive, high-performance memory operations.  

\paragraph{Type-Aware Transformation and Loading Mechanism}  

During memory scheduling, \texttt{MemScheduler} analyzes task semantics, window size, and resource constraints to determine the best-fit memory type.  
Stable, frequently accessed content is transformed into \textbf{Activation Memory} for KV caching, minimizing prefill latency.  
Abstract rules and reusable patterns are encoded as \textbf{Parameter Memory}—e.g., via distillation or adapters embedded into model weights.  
Time-sensitive or session-specific knowledge is preserved as \textbf{Plaintext Memory}, inserted into the prompt as raw text.  
Adaptive triggers guide the loading process.  
For coherence-heavy tasks like multi-turn dialogue, the scheduler favors KV-cache recall.  
For procedural or expert-driven flows, parametric modules take precedence.  
For on-demand factual queries, plain memory is retrieved and contextualized.  
All decisions are logged to \texttt{MemCube} and coordinated with \texttt{MemOperator}'s memory structure to maintain traceability and interpretability.  

\paragraph{Cross-Type Conversion and Migration}  
To maintain long-term performance and adaptive memory utilization, \texttt{MemScheduler} supports cross-type memory migration.  
For example, plain memories frequently recalled across sessions may be promoted to Activation Memory (KV cache).  
Stable templates used repeatedly can be distilled into parameter Memory.  
Conversely, underutilized KV entries may be downgraded to Plain Memory and archived to cold storage.  
This type-shifting mechanism ensures memory units evolve toward their optimal invocation form while conserving system resources.  

\paragraph{Execution Path Integration and Governance}  
\texttt{MemScheduler} integrates upstream with \texttt{MemReader} and the \texttt{Memory API} to parse structured calls and semantic goals.  
Downstream, it collaborates with model execution paths to determine how and where to inject memory.  
Scheduling logic is optimized in real time, guided by task type, model load, cache hit rates, and access history.  
All dispatch actions are governed by \texttt{MemGovernance}, which enforces user-role boundaries, rate limits, and lifecycle policies.  
This ensures proper memory isolation and secure usage across users, models, and tasks, while maintaining an auditable record of every memory interaction.  

\subsubsection{MemLifecycle}

In \textsc{MemOS}, each memory object is treated as a dynamic entity with evolving states, managed centrally by the \texttt{MemLifecycle} module.
The system models memory as a finite state machine, cycling through four key states: Generated, Activated, Merged, and Archived.
This framework supports semantic evolution, dynamic memory management, and stable, controlled resource scheduling at the storage layer.

\paragraph{State Modeling and Evolution Logic}  
State transitions are triggered by a combination of system policies and user actions.
For instance, in a smart meeting assistant, an auto-generated summary is initially labeled as “Generated”.
If that summary is later referenced in a follow-up task—like agenda tracking or meeting comparison—it transitions into the “Activated” state.
When the user adds supplementary data, or the system detects semantic overlap with historical memory, these entries are consolidated into a new version and marked as “Merged”.
If a memory is no longer accessed for a prolonged period, it is demoted to the “Archived” state and moved to cold storage.
Transitions can be explicitly initiated by user actions, or implicitly driven by system heuristics such as recency, contextual salience, or successful merge events.

\paragraph{Time Machine and Freezing Mechanism}  
To ensure long-term consistency and recovery, \textsc{MemOS} offers a “Time Machine” capability that snapshots memory states and supports historical rollbacks.
Users or developers can invoke this feature to restore an archived or merged memory back to a specific version, re-enabling its use in inference and context injection.
This is critical for scenarios such as detecting model forgetting, handling user retractions, or conducting counterfactual simulations.
In a policy collaboration platform, a user might unarchive an old clause to perform “what-if” simulations, without impacting the canonical frozen version and its audit trail.
\textsc{MemOS} also supports a “Frozen” state for critical memories—like legal agreements or standard guidelines—where updates are disabled and full modification histories are retained for auditing, compliance, or education.

\paragraph{Scheduling and Storage Integration Strategy}  
Lifecycle states directly influence scheduling priority and storage allocation strategies.
“Activated” memories are preferentially cached in local memory or fast-access \texttt{MemoryCube} instances for low-latency retrieval.
“Archived” or “Frozen” memories are offloaded to \texttt{MemVault}, a cold storage layer optimized for durability over speed.
Based on lifecycle rules, the system can batch-trigger operations like cleanup, compression, or migration to balance call availability with efficient resource usage.

\subsection{Infrastructure Layer}

\subsubsection{MemGovernance}
\texttt{MemGovernance} is the core module in \textsc{MemOS} responsible for memory access control, compliance enforcement, and auditability.
As memory systems evolve toward multi-user collaboration and long-horizon task reasoning, \texttt{MemGovernance} ensures that memory remains secure, interpretable, and controllable throughout its sharing, transfer, and inference processes.

It establishes a ternary permission model involving the user identity, the memory object, and the calling context, supporting private, shared, and read-only access policies.
Each memory request undergoes identity authentication and contextual validation to prevent unauthorized access.
For example, in clinical settings, only physicians may access a patient’s diagnostic records; in enterprise systems, only authorized managers can retrieve archived policy documents.

It manages memory lifecycle policies such as time-to-live (TTL) enforcement and access-frequency-based garbage collection or archiving of inactive items.
It also tracks memory usage heat to monitor high-traffic memory segments.
Its privacy control subsystem includes sensitive content detection, automatic redaction, and access logging to ensure personal and behavioral data remain secure.

All memory objects carry full provenance metadata, including creation source, invocation lineage, and mutation logs.
Generated content can be watermarked semantically and tagged with behavioral fingerprints, allowing attribution and copyright tracking in multi-platform scenarios.

The module also exposes audit interfaces for integration with enterprise compliance systems, supporting export of access logs and permission revision reports.
These features support regulatory compliance in high-stakes environments such as healthcare and finance.

\subsubsection{MemVault}
\texttt{MemVault} is the central memory storage and routing infrastructure in \textsc{MemOS}, responsible for managing and serving diverse categories of memory.
Memory is organized into namespaces such as user-private stores, expert knowledge bases, industry-shared repositories, contextual memory pools, and pipeline-aligned caches.
Each is assigned a dedicated namespace and path structure to support efficient lookup and access control.

To support heterogeneous backends, \texttt{MemVault} interfaces with vector stores, relational databases, and blob storage through a unified \texttt{MemoryAdapter} abstraction.
This allows API-level consistency for querying, writing, and syncing memory regardless of backend heterogeneity.
Stores may be configured as read-only caches or write-enabled repositories, depending on latency or learning objectives.

At runtime, \texttt{MemVault} works in concert with \texttt{MemScheduler} and \texttt{MemLifecycle} to dynamically load memory based on access history, contextual relevance, and memory state.
It supports tag-based, semantic, and full-text loading patterns, and triggers migration for hot memory to fast storage or cold data to archival zones.
This architecture is vital for multi-model collaboration, domain-level knowledge fusion, and consistency in multi-turn dialogue—forming the knowledge backbone for scalable intelligent systems.

\subsubsection{MemLoader \& MemDumper}
\texttt{MemLoader} and \texttt{MemDumper} form a bi-directional channel for memory migration across platforms in \textsc{MemOS}.
They support injection, export, and synchronization of structured units like \texttt{MemoryCube}.
This capability is essential for system handover, edge-cloud integration, and knowledge continuity across distributed agents.

During ingestion, \texttt{MemLoader} accepts memory from local caches, third-party systems, or archives and maps it to target stores.
It auto-fills provenance metadata, tagging, and lifecycle status to ensure governance readiness.

\texttt{MemDumper} exports selected memory in portable formats with permission metadata, redacted fields, and access logs.
Both components support periodic and event-driven updates, such as automatic export upon tag activation.
The migration process is governed by \texttt{MemGovernance} to validate policies, trace operations, and isolate sensitive data.
For instance, a mobile device may export patient interaction logs to the cloud, which remote agents later load to preserve task context.

\subsubsection{MemStore}
\texttt{MemStore} is the open-access interface in \textsc{MemOS} that enables controlled publishing, subscription, and distribution of memory units.
It supports memory exchange between models, institutions, and even industry-wide networks.

Users may declare memory as publishable and define visibility, usage conditions, and access control rules.
Each shared unit carries unique IDs and provenance metadata; \texttt{MemGovernance} ensures masking, watermarking, and policy validation during dissemination.

\texttt{MemStore} enables both push and pull models of memory exchange.
Consumers can define subscriptions using tags or semantic filters, and the system delivers matched updates proactively.
Licensed memory assets can enforce contract-bound access frequencies and expiry policies.
All access is logged with invocation traces to support audit and accountability.

For example, a hospital may publish de-identified diagnostic records for remote triage agents, with every call validated for context and provenance.

\section{Evaluation}

To systematically evaluate the capabilities of \textsc{MemOS}, we conduct both holistic and component-level experiments.
We begin by benchmarking the full system on the LoCoMo~\cite{maharana2024evaluating}, LongMemEval~\cite{wu2024longmemeval}, PreFEval~\cite{zhao2025llms} and PersonaMem~\cite{jiang2025know} benchmark suite to assess its performance in memory-intensive reasoning and personalization tasks, comparing against several state-of-the-art baselines.
In addition, we present targeted evaluations of key architectural subsystems, including multi-perspective memory organization, hybrid semantic retrieval, task-aligned scheduling, and KV-based activation memory injection. These experiments assess the individual effectiveness of each component and its contribution to overall system performance.

\subsection{End-to-End Evaluation on Long Context Memory}\label{sec:long_context_memory}

To evaluate long-context and multi-session capabilities, we test \textsc{MemOS} on the LoCoMo~\cite{maharana2024evaluating} and LongMemEval~\cite{wu2024longmemeval} benchmarks against a diverse set of strong baselines, each representing a distinct memory design paradigm.

Specifically, \textit{MIRIX}~\cite{wang2025mirix} manages memory via six specialized components (Core, Episodic, Semantic, Procedural, Resource, and Knowledge Vault);
\textit{Mem0}~\cite{chhikara2025mem0} implements slot-based long-term memory with top-k semantic search;
\textit{Zep}~\cite{rasmussen2025zep} integrates time-aware knowledge graphs with structured query resolution;
\textit{Memobase} prioritizes the balance between performance, cost, and latency in long-term user memory;
\textit{Supermemory} employs a Dynamic Knowledge Graph to map relationships between memories, enabling semantic understanding as information evolves;
and \textit{MemU} processes multi-modal inputs by extracting and summarizing them into structured memory files.
To ensure architectural parity, all methods are implemented over the same LLM backbone (GPT-4o-mini~\cite{achiam_2023_gpt4}).

All experiments are conducted on an 80GB H800 GPU under identical hardware and software configurations. To ensure a fair and optimized comparison, the configuration for each method is selected based on its best validation performance. Note that for \textit{MIRIX}, token consumption data is omitted from the results due to system limitations.

\input{tab/locomo}

As shown in Table~\ref{tab:new_locomo}, \textsc{MemOS} achieves the best average performance across nearly all task categories on the LoCoMo benchmark.
Across all sub-tasks in LoCoMo, \textsc{MemOS} consistently ranks among the top performers, maintaining first or second place in every category. It demonstrates clear advantages in single-hop and multi-hop tasks, where long-range memory and contextual integration are especially critical. Beyond LLM-judge scores, \textsc{MemOS} also delivers strong generation quality in terms of F1 score while maintaining reasonable context length control, implying that its high LLM-judge scores are not a result of retrieval token overflow.

\input{tab/longmemeval}

Similarly, for LongMemEval (Table~\ref{tab:longmemeval}), \textsc{MemOS} achieves the best average performance across almost all task categories.
Across all sub-tasks in the benchmark, \textsc{MemOS} ranks among the top performers, placing first or second in every category except knowledge updates, and securing the best overall performance on average.

\subsection{End-to-End Evaluation on Personalization and Preference Understanding}

To evaluate personalization and preference understanding, we test \textsc{MemOS} on the PreFEval~\cite{zhao2025llms} and PersonaMem~\cite{jiang2025know} benchmarks against the same baselines detailed in Section \ref{sec:long_context_memory}. We use the identical hardware and software configurations as described in Section \ref{sec:long_context_memory}. For a fair and optimized comparison, the configuration for each method is selected based on its best validation performance.

\input{tab/perfeval}

As shown in Table \ref{tab:perfeval}, \textsc{MemOS} not only achieved the best Personalized Response performance in both scenarios (with 0 turns and with 10 irrelevant turns) but also recorded the lowest preference unaware error. Importantly, it maintained acceptable context length control and low overall error rates. This demonstrates \textsc{MemOS}'s stable recognition of user preferences and robustness in long-term memory, highlighting its unique value in understanding user intent and providing highly personalized experiences.

\input{tab/personamem}

Similarly, for the PersonaMem benchmark (Table \ref{tab:personamem}), \textsc{MemOS} achieved the best precision while maintaining acceptable context length control, further validating its superior capability in handling dynamic user profiles and preferences.

\subsection{Evaluation of chunk sizes and Top-K selection}

\begin{figure}[ht]
    \centering
    \includegraphics[width=\linewidth]{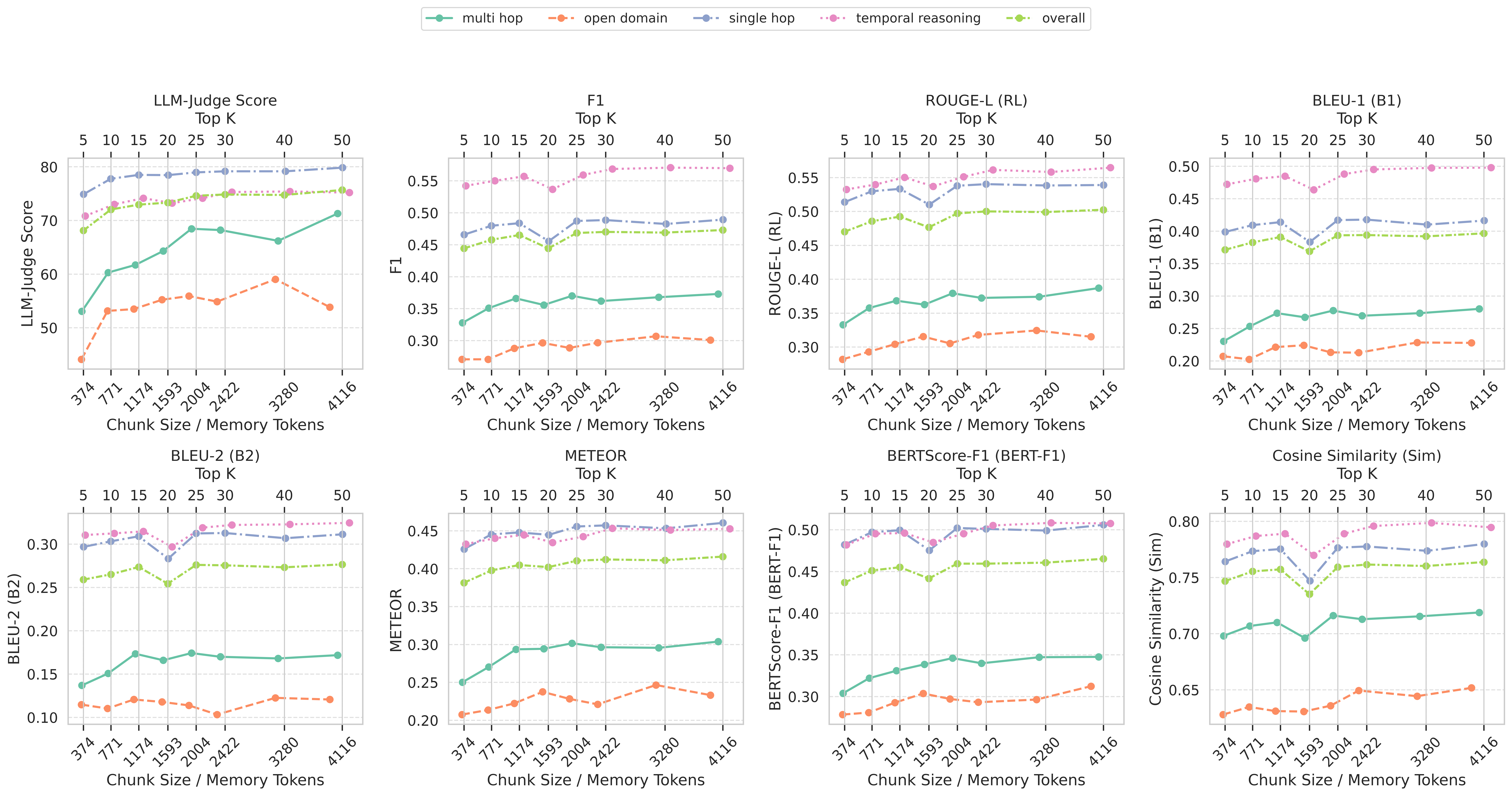}
    \caption{
    \textbf{Performance trends of \textsc{MemOS} across memory configurations.} We vary the number of retrieved memory chunks (Top-K, upper x-axis) and chunk size (lower x-axis, total memory tokens), and report performance on the LoCoMo benchmark across multiple metrics and task types. MemOS consistently maintains top-tier performance as memory capacity increases, with clear gains on multi-hop and temporal reasoning tasks. Cosine similarity results indicate stable semantic alignment throughout.
    }
    \label{fig:ablation_result}
\end{figure}

To better understand the impact of memory configuration, we conduct an ablation study by varying chunk sizes and Top-K retrieval depth. 
As shown in Figure~\ref{fig:ablation_result}, \textsc{MemOS} demonstrates stable and strong performance across all LoCoMo sub-tasks, with performance steadily improving as memory capacity increases—particularly for multi-hop and temporal reasoning tasks that demand long-range retrieval and contextual integration.
In addition to higher LLM-Judge scores, generation metrics such as F1, ROUGE-L, and BLEU also benefit from memory expansion. Cosine similarity remains consistently high, indicating stable semantic alignment even with deeper retrieval.
 
These results collectively validate the effectiveness of \textsc{MemOS}'s architectural innovations—particularly its hybrid semantic retrieval and memory-centric design—which enable accurate, fluent, and contextually aligned responses under long-horizon constraints.

\subsection{Evaluation of Memory Retrieval Robustness}

\input{tab/latency}

We conduct a focused evaluation to analyze the efficiency and effectiveness of memory retrieval via network API. As shown in Table~\ref{tab:time_latency}, we compare latency and success percentage for various API-based baselines under different Queries Per Second (QPS) request pressures. The baselines include \textit{Mem0}, \textit{Memobase}, \textit{Supermemory}, \textit{MemU}, and \textit{Zep}, which are introduced in Section \ref{sec:long_context_memory}. The request contents are randomly sampled from the LoCoMo benchmark.

To test robustness, the metrics reported include the P99, P90, and mean latency, as well as the success percentage rate for both memory insertion (add) and retrieval (search) operations.

Our results show that \textsc{MemOS} exhibited the highest robustness with a 100\% success percentage, maintaining the lowest latency across nearly all metrics regardless of the QPS pressure. Remarkably, \textsc{MemOS} achieved a 100\% success rate and maintained reasonable latency even under 100 QPS. This demonstrates that \textsc{MemOS}'s hybrid semantic organization and activation-based memory loading can achieve superior and highly stable performance across a wide range of QPS pressure.

\subsection{Evaluation of KV-Based Memory Acceleration}

To evaluate the effectiveness of KV-form memory acceleration within \textsc{MemOS}, we design a controlled experiment simulating realistic memory reuse scenarios.

During typical usage, the \textit{MemScheduler} module in \textsc{MemOS} continuously monitors model interactions and automatically identifies the most frequently accessed and semantically stable plaintext memory entries. These entries are then converted into \textbf{activation memory}—a KV-format structure injected into the model's attention cache and proactively transferred to GPU memory for low-latency reuse.

Our evaluation assumes this realistic deployment: memory has already been preprocessed and cached on GPU in KV format, avoiding the need for repeated prompt encoding.

We compare two memory usage strategies: prompt-based memory injection, where memory entries are prepended to the input sequence, and KV-cache injection, where memory is injected directly as key-value pairs into the model's attention mechanism.

To simulate realistic inference conditions, we evaluate across three context lengths—short (583 tokens), medium (2773 tokens), and long (6064 tokens)—as well as three query types of increasing length and complexity: short (167 tokens), medium (302.7 tokens), and long (952.7 tokens).
All experiments are conducted using the HuggingFace \texttt{transformers} library, running on a single NVIDIA H800 GPU with 80GB of memory under consistent system settings.

We report four metrics as shown in Table~\ref{tab:ttft-performance}. “Build” time refers to the preprocessing duration needed to convert memory into KV format. “KV TTFT” denotes the first-token latency under KV-based memory injection, while “Dir TTFT” indicates the latency under prompt-based injection. “Speedup” reflects the relative latency reduction achieved by KV injection compared to direct prompt injection.

\input{tab/kvtest}

\begin{figure}[h]
    \centering
    \includegraphics[width=0.85\textwidth]{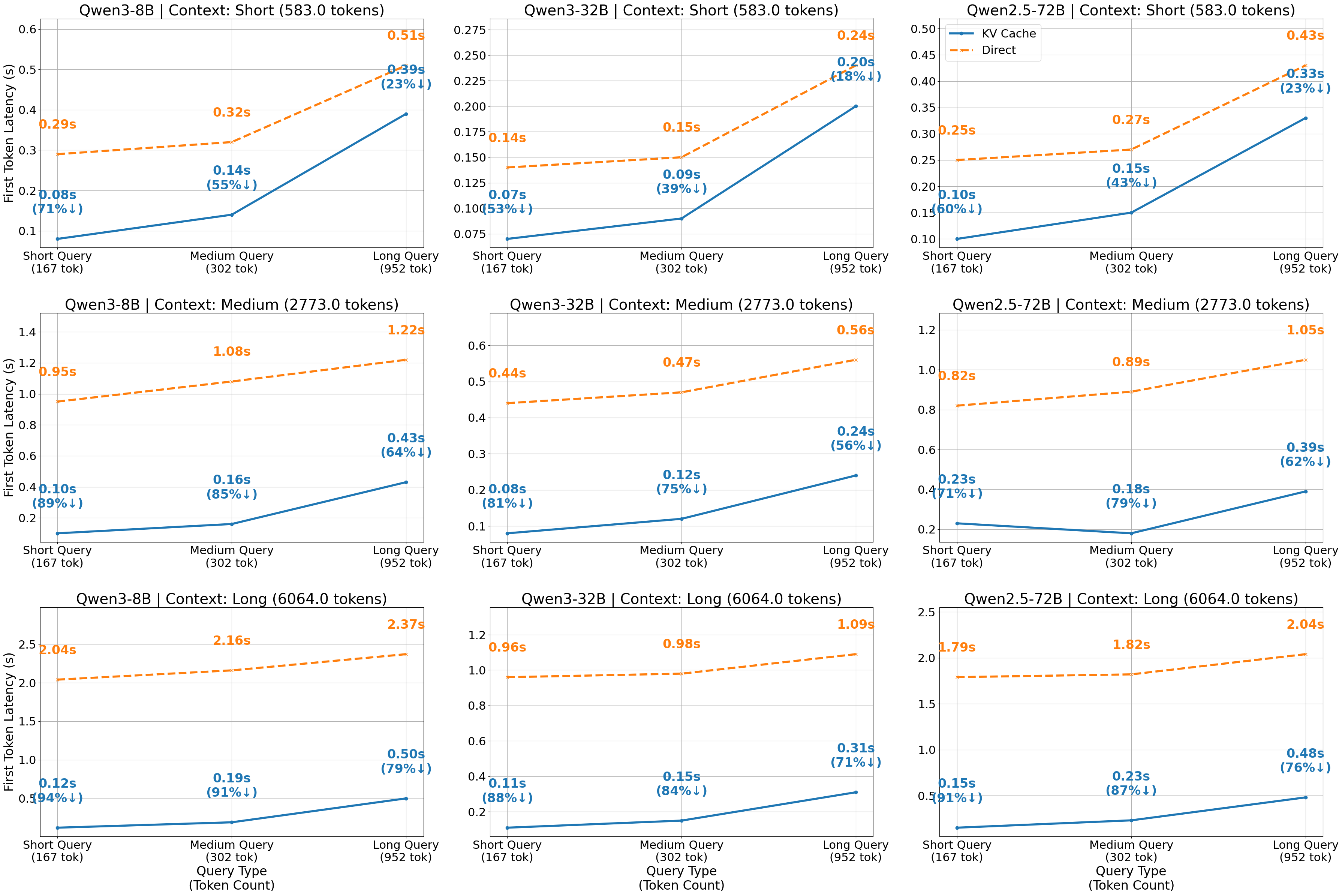}
    \caption{Time to First Token (TTFT) comparison across models, context lengths, and query lengths. KV-based memory injection consistently achieves lower latency with identical output.}
    \label{fig:ttft-performance}
\end{figure}

The results (Table~\ref{tab:ttft-performance} and Figure~\ref{fig:ttft-performance}) confirm that KV-based memory injection yields substantial TTFT reduction across all models and configurations. The output sequences remain identical under both methods, validating their semantic equivalence. Acceleration is especially significant for larger models and longer contexts—for instance, Qwen2.5-72B achieves a 91.4\% reduction in TTFT under long-context, short-query conditions. These findings highlight KV memory as a practical and effective technique for low-latency execution in memory-augmented language models.

\section{\textsc{MemOS} for Architecture Innovation and Applications}

\subsection{Architectural Innovations Enabled by \textsc{MemOS}}

\textsc{MemOS} treats memory as a first-class system resource, enabling unified lifecycle management and orchestration of memory in multiple forms.
This abstraction supports architectural innovations that focus on memory-driven modules and services, facilitating the modularization and reusability of knowledge assets.

\subsubsection{Paid Memory as Modular Installables (User-Facing Paradigm)}

\textsc{MemOS} is designed around a memory-centric architecture, offering modularized and assetized memory interfaces that allow knowledge to be uploaded, mounted, and invoked like a digital resource.
Under this paradigm, memory is no longer bound to training pipelines or development workflows but becomes a composable and user-controllable intelligence unit.

Concretely, domain experts can publish structured experiential memories via \texttt{MemStore}, akin to publishing a knowledge plugin or an expert tip.
Consumers—students, enterprise agents, or assistant models—can install these memories using a standardized loading interface, subject to permission control.
This entire flow abstracts away the need for understanding the underlying model architecture or performing manual alignment.
It drastically reduces the barrier to memory usage and makes memory-driven intelligence available beyond developers and platform operators.

For instance, a medical student in clinical rotation may wish to study how to manage a rare autoimmune condition.
An experienced physician can encapsulate diagnostic heuristics, questioning paths, and typical case patterns into a structured memory and upload it to \texttt{MemStore}.
The student can then search, install, and invoke this memory locally via their assistant model.
This bypasses the need for building formal ontologies or coordinating structured knowledge base design, as is common in traditional clinical AI.

\textsc{MemOS} encapsulates this process as a standardized "Memory-as-a-Service" capability, greatly expanding the accessibility and reusability of expert knowledge.
Furthermore, \texttt{MemGovernance}, the dedicated control module in \textsc{MemOS}, offers full-spectrum privacy and access control for memory assets.
It enables memory providers to define custom access conditions over their published content.
For example, a medical expert may restrict installation rights to users who have completed a micropayment, enabling a form of licensed intelligence delivery.

\subsubsection{Painless Memory Management (Task-Oriented Paradigm)}

\textsc{MemOS} abstracts memory as a universal, long-lived, and shareable infrastructure resource, architecturally analogous to storage subsystems in traditional operating systems.
This design elevates memory from a model-embedded utility to a first-class system-level asset with its own lifecycle and invocation semantics.

Unlike conventional transient memory techniques limited to context windows or parameter embeddings,
\textsc{MemOS} offers standardized memory interfaces, a unified access protocol, and structured persistence formats.
This enables runtime tasks to flexibly read, write, mount, fuse, or replace memory blocks on demand,
without requiring manual state tracking or architectural alignment.

Neither users nor developers need to handle low-level vector indexing, KV-caching, or context orchestration logic.
Instead, they can access and update memory seamlessly through task-level \texttt{Memory API} calls.
This infrastructure-level abstraction proves especially valuable in multi-stage, long-horizon, and evolving tasks.

For example, in an intelligent legal assistant system, a user may complete a corporate contract review task in distinct phases:
the first phase may focus on structural layout and terminological consistency;
the second phase may highlight risky clauses and compare precedent cases;
and the final phase may involve checking compliance against current regulations.
\textsc{MemOS} dynamically loads the appropriate memory sets at each stage (e.g., “Contract Template Memory”, “Risk Clause Case Logs”, “Recent Regulation Digest”),
and performs hot-swapping and cache eviction as task contexts evolve.
Throughout the task lifecycle, the user need not explicitly manage memory policies;
the system automatically schedules the relevant memory assets based on context semantics,
delivering a “memory-as-resource, use-on-demand” intelligent task execution experience.

\subsection{\textsc{MemOS} Application Scenarios}

\subsubsection{Supporting Multi-Turn Dialogue and Cross-Task Continuity}

Real-world interactions rarely reveal user intent in a single turn; instead, goals are refined progressively over multiple exchanges. However, traditional LLMs rely on static context windows, making it difficult to retain key semantic states across turns, resulting in “memory loss” between dialogue rounds.

For instance, in a procurement negotiation task, a user might set a budget cap of ¥300,000 in round 5, later revise product preferences in round 12 to prioritize domestic alternatives, yet by round 15, the model reverts to recommending high-priced imports based on earlier defaults.

\textsc{MemOS} addresses this at the system level by extracting salient elements (e.g., budget, preferences, delivery constraints) after each user input and encoding them into structured “conversation memory units.” These are linked to the ongoing task’s long-term memory path via \texttt{MemLink}.

During inference, \texttt{MemScheduler} retrieves relevant historical fragments based on current context and integrates them into the active reasoning path. This ensures continuity of semantic state and prevents logic drift due to “context sliding.”

Furthermore, \textsc{MemOS} supports cross-task memory reuse to enable dialogue continuity and state persistence. For example, after completing an auto-form-filling task, the system retains memory of ID details or user habits. When the user later initiates a “visa application” task, \textsc{MemOS} recalls the previously stored data (e.g., from “passport issuance”), enabling seamless state transition across tasks.

\subsubsection{Supporting Knowledge Evolution and Continuous Update}

Modern knowledge is dynamic, yet LLMs are generally trained once with static datasets. Updating their internal knowledge either requires expensive fine-tuning or introduces risks like catastrophic forgetting. Even RAG approaches lack lifecycle, version, or governance mechanisms—leading to fragmented, unverifiable external knowledge.

\textsc{MemOS} redefines knowledge as dynamic, lifecycle-governed memory. Each memory unit evolves independently, with defined stages for generation, replacement, fusion, and deprecation. The system schedules updates based on usage frequency, contextual alignment, and semantic overlap.

For example, when updated clinical guidelines are published, medical authorities can release them as explicit memory blocks via \texttt{MemStore}. \textsc{MemOS} tags them as “trusted sources,” compares them with older versions, and suggests updates to users.

At inference time, \texttt{MemScheduler} prioritizes trusted and active versions, while obsolete entries are archived. This allows the model to remain up-to-date without retraining or harming prior knowledge structures.

\textsc{MemOS} also supports personalized knowledge development. For instance, a cancer specialist may iteratively add interpretations and heuristics to drug usage. Over time, these refinements are integrated into their personal memory path, coexisting with official guidelines and selected based on task context.

\subsubsection{Enabling Personalization and Multi-Role Modeling}

LLMs today often operate statelessly across users and roles, unable to remember stylistic preferences or distinguish between user roles in complex settings. As a result, users must re-specify information every time, and models struggle to maintain consistent identity or behavior.

\textsc{MemOS} provides system-level support for identity-aware memory and role-based behavior. Each user identity is associated with dedicated memory spaces, and multiple roles can coexist under one account.

For example, a user may interact as both a “parent” managing home tasks and a “manager” handling contracts. \textsc{MemOS} keeps memory streams separate and dynamically loads the appropriate persona during inference.

In addition, long-term interaction patterns are encoded into “personal memory units” capturing language tone, response preferences, or value leanings. These units are incorporated into inference, yielding a personalized and coherent AI behavior.

In enterprise contexts, \textsc{MemOS} allows deployment of predefined role templates with task scopes, permission controls, and memory sync strategies. For example, an organization may define roles for analysts, assistants, and project leads, each with distinct memory access and agent behavior.

\subsubsection{Enabling Cross-Platform Memory Migration}

In a world of multi-device, multi-agent environments, valuable user-model memories often become locked within individual platforms, creating “memory silos” that break continuity and fragment knowledge accumulation.

\textsc{MemOS} resolves this through standardized memory representations, encryption, and platform-agnostic mount protocols. All memory blocks are portable across environments—from mobile to cloud to enterprise infrastructure.

For example, a user’s “family travel preference” memory built via mobile assistant—including flight timing, hotel type, and budget—can be selectively migrated to a corporate travel planning agent on desktop, enabling consistent and efficient decision-making.

By breaking the memory silo, \textsc{MemOS} transforms memory from a private asset embedded in a single model to a distributed, governable, and reusable intelligence layer across platforms.

\section{Conclusion}
In this work, we introduce a memory operating system designed for Large Language Models, aimed at collaboratively building foundational memory infrastructure for next-generation LLM applications.

\textsc{MemOS} provides a unified abstraction and integrated management framework for heterogeneous memory types, including parameter memory, activation memory, and explicit plaintext memory. We propose a standardized memory unit, \texttt{MemCube}, and implement key modules for scheduling, lifecycle management, structured storage, and transparent augmentation. These components collectively enhance reasoning coherence, adaptability, and system scalability in LLMs.

Building on this foundation, we envision a future intelligent ecosystem centered on modular memory resources and supported by a decentralized memory marketplace. This paradigm shift enables the creation of next-generation AI systems capable of continual learning and long-term evolution.

Looking ahead, we plan to explore the following directions:
\begin{itemize}
\item \textbf{Cross-LLM Memory Sharing}: Enable interoperability and module reuse across different foundation models by sharing parametric and activation memories. To support consistent semantics and secure exchange, we plan to extend the \textbf{Memory Interchange Protocol (MIP)} to define standard formats, compatibility rules, and trust mechanisms for cross-model/app memory transmission—facilitating collaborative knowledge transfer among agents.
\item \textbf{Self-Evolving MemBlocks}: Develop memory units capable of self-optimization, reconstruction, and evolution based on usage feedback, reducing the need for manual maintenance and supervision.
\item \textbf{Scalable Memory Marketplace}: Establish decentralized mechanisms for memory exchange, supporting asset-level transactions, collaborative updates, and distributed evolution to foster a sustainable AI ecosystem.
\end{itemize}

Overall, with the introduction of \textsc{MemOS}, we aim to transform LLMs from closed, static generation systems to continuously evolving intelligent agents equipped with long-term memory, integrated knowledge, and behavioral plasticity. \textsc{MemOS} not only addresses critical architectural limitations in current models but also lays the groundwork for cross-task, cross-platform, and multi-agent collaborative intelligence. Building on prior work demonstrating the potential of explicit memory and hierarchical memory representations in LLMs~\cite{memory3_Yang_2024}, we look forward to advancing the frontiers of \textsc{MemOS} in collaboration with the community, making memory a first-class computational resource in the age of general-purpose AI.

\bibliographystyle{unsrt}
\bibliography{main}

\end{document}

%% file: tab/references.tex
\begin{table}[htbp]
\centering
\caption{Classification of Memory Types, Mechanisms, and Example References}
\label{tab:memory_classification}
\renewcommand{\arraystretch}{1.4}
\footnotesize
\begin{tabular}{m{0.08\textwidth} m{0.12\textwidth} m{0.23\textwidth} m{0.46\textwidth}}
\hline
\textbf{Timescale} & \textbf{Consciousness} & \textbf{Mechanism} & \textbf{Example References} \\ \hline

\multirow{4}{*}[-4ex]{\makecell[l]{Short-term}} 
& \multirow{1}{*}{\makecell[l]{Explicit}} 
& Prompt-Based Context 
& GPT-2 \cite{radford_gpt2_nodate}, GPT-3 \cite{brown_gpt3_2020}, Prefix-Tuning \cite{li_prefix-tuning_2021}, Prompt-Tuning \cite{lester_prompt-tuning_2021}, P-Tuning \cite{liu_ptuning_2023, liu_ptuningv2_2022}, InstructGPT \cite{ouyang_instructgpt_2022} \\ \cline{2-4}

& \multirow{3}{*}[-2ex]{\makecell[l]{Implicit}} 
& Key-Value Cache Mechanism 
& vLLM \cite{kwon_vLLM_2023}, StreamingLLM\cite{xiao_streamingllm_2023}, H2O\cite{zhang_h2o_2023}, LESS \cite{dong_less_2024}, KVQuant \cite{hooper_kvquant_2024}, RetrievalAttention \cite{liu_retrievalattention_2024}, Memory\textsuperscript{3} \cite{memory3_Yang_2024} \\ \cline{3-4}

& 
& Hidden State Steering 
& Steer \cite{subramani_steerno1_2022}, ICV \cite{liu_ICV_2024}, ActAdd \cite{turner_ActAdd_2024}, StyleVec \cite{konen_stylevec_2024}, CAA \cite{rimsky_CAA_2024}, FreeCtrl \cite{feng_freectrl_2024}, EasyEdit2 \cite{xu_easyedit2_2025} \\ \cline{3-4}

& 
& Activation Circuit Modulation 
& SAC \cite{xiao_sac_2024}, DESTEIN \cite{li_destein_2024}, LM-Steer \cite{han_lmsteer_2024} \\ \hline

\multirow{4}{*}[-6ex]{\makecell[l]{Long-term}} 
& \multirow{1}{*}[-2ex]{\makecell[l]{Explicit}} 
& Non-parametric Retrieval-Augmented Generation 
& kNN-LMs \cite{khandelwal_knnlm_2019,pozzobon_goodtriever_2023}, MEMWALKER \cite{DBLP:journals/corr/abs-2310-05029}, Graph RAG \cite{DBLP:journals/corr/abs-2404-16130}, LightRAG \cite{DBLP:journals/corr/abs-2410-05779}, NodeRAG \cite{xu2025noderagstructuringgraphbasedrag,yang2025heteragheterogeneousretrievalaugmentedgeneration}, HyperGraphRAG \cite{DBLP:journals/corr/abs-2503-21322}, HippoRAG \cite{DBLP:conf/nips/GutierrezS0Y024,DBLP:journals/corr/abs-2502-14802}, PGRAG\cite{DBLP:journals/corr/abs-2405-16933}, Zep \cite{DBLP:journals/corr/abs-2501-13956}, A-MEM~\cite{DBLP:journals/corr/abs-2502-12110}, Mem0\cite{chhikara2025mem0buildingproductionreadyai} \\ \cline{2-4}

& \multirow{3}{*}[-4ex]{\makecell[l]{Implicit}} 
& Parametric Knowledge 
& BERT \cite{devlin_bert_2019}, RLHF \cite{bai_rlhf_2022}, CTRL \cite{keskar_ctrl_2019}, SLayer \cite{chen_slayer_2024} \\ \cline{3-4}

& 
& Modular Parameter Adaptation 
& LoRA \cite{hu_lora_2021}, PRAG \cite{su_PRAG_2025}, DyPRAG \cite{tan_DyPRAG_2025}, SERAC \cite{mitchell_serac_2022}, CaliNet \cite{dong_CaliNet_2022}, DPM \cite{cheng_DPM_2023}, GRACE \cite{hartvigsen_grace_2023} \\ \cline{3-4}

& 
& Parametric Memory Editing 
& ROME \cite{meng_rome_2023}, MEMIT \cite{meng_memit_2023}, AlphaEdit \cite{fang_alphaedit_2025}, AnyEdit \cite{jiang_anyedit_2025}, EasyEdit \cite{zhang_easyedit_2024}, AdaPLE \cite{li_AdaPLE_2024}, MEMAT \cite{tamayo_memat_2024} \\ \hline

\end{tabular}
\end{table}

%% file: tab/osmapping.tex
\begin{table}[h]
\centering
\footnotesize
\caption{Mapping of Traditional OS Components to MemOS Modules}
\label{table:mapping}
\resizebox{\textwidth}{!}{
\begin{tabular}{llll}
\toprule
Layer & OS Component & MemOS Module & Role \\
\midrule
\multicolumn{4}{c}{\textit{Core Operation Layer}} \\
\midrule
Parameter Memory  & Registers / Microcode & Parameter Memory & Long-term ability \\
Activation Memory & Cache                 & Activation Memory & Fast working state \\
Plaintext Memory  & I/O Buffer            & Plaintext Memory  & External episodes \\
\midrule
\multicolumn{4}{c}{\textit{Management Layer}} \\
\midrule
Scheduling        & Scheduler      & MemScheduler         & Prioritise ops \\
Persistent Store  & File System    & MemVault             & Versioned store \\
System Interface  & System Call    & Memory API           & Unified access \\
Backend Driver    & Device Driver  & MemLoader / Dumper   & Move memories \\
Package Deploy    & Package Manager& MemStore             & Share bundles \\
\midrule
\multicolumn{4}{c}{\textit{Governance \& Observability}} \\
\midrule
Auth / ACLs       & Auth Module, ACLs & MemGovernance      & Access control \\
Logging           & Syslog            & Audit Log          & Audit trail \\
Fault Handling    & Excp. Handler     & Error Recovery     & Error recover \\
\bottomrule
\end{tabular}}
\end{table}

%% file: tab/locomo.tex
\begin{table}[htbp]
\centering
\scriptsize
\renewcommand{\arraystretch}{1.3}
\caption{
We evaluated our \textsc{MemOS-1031} model against various baselines on the LoCoMo benchmark, all using the same GPT-4o-mini foundation LLM. 
We report the performance based on LLM Judge Scores across five major tasks (single-hop, multi-hop, temporal reasoning, open-domain, and overall), alongside the overall F1 score and the context tokens used by each method. 
The results demonstrate that \textsc{MemOS-1031} consistently outperformed baseline methods in almost all scenarios, all while maintaining acceptable context length and a reasonable F1 score.
}
\resizebox{\textwidth}{!}{
\begin{tabular}{l|ccccc|cc}
\hline
Method                                          & Tokens& Single-hop $\uparrow$     & Multi-hop $\uparrow$  & Temporal Reasoning $\uparrow$ & Open-domain $\uparrow$    & Overall $\uparrow$    & Overall F1 $\uparrow$ \\\hline
MIRIX                                           & -     & 68.22                     & 54.26                     & 68.54                             & 46.88                         & 64.33                     & 28.10 \\
Mem0                                            & 1172  & 73.33                     & 58.75                     & 52.34                             & 45.83                         & 64.57                     & 43.46 \\
Zep                                             & 2701  & 66.23                     & 52.12                     & 54.82                             & 33.33                         & 59.22                     & 41.23 \\
Memobase                                        & 2102  & 73.12                     & 64.65                     & \textbf{81.20}                             & 53.12                         & 72.01                     & \textbf{50.18} \\
MemU                                            & 617   & 66.34                     & 63.12                     & 27.10                             & 50.01                         & 56.55                     & 35.15 \\
Supermemory                                     & 500   & 67.30                     & 51.12                     & 31.77                             & 42.67                         & 55.34                     & 34.87 \\
\rowcolor[HTML]{EFEFEF} \textsc{MemOS-1031}     & 1589  & \textbf{81.09}            & \textbf{67.49}            & 75.18                    & \textbf{55.90}                & \textbf{75.80}            & 45.27 \\\hline
\end{tabular}
}
\label{tab:new_locomo}
\end{table}

%% file: tab/longmemeval.tex
\begin{table}[ht]
\centering
\scriptsize
\renewcommand{\arraystretch}{1.3}
\caption{
We evaluated our \textsc{MemOS-1031} model against various baselines on the LongMemEval benchmark, all using the same GPT-4o-mini foundation LLM. 
We report the performance percentage across six key scenarios (single-session preference, single-session assistant, temporal reasoning, multi-session, knowledge update, and single-session user), along with the context tokens used by each method. 
The results demonstrate that \textsc{MemOS-1031} achieved the best precision in almost all scenarios while maintaining acceptable control over context length.
}
\resizebox{\textwidth}{!}{
\begin{tabular}{l|ccccccc|c}
\hline
method      & \makecell{content \\ tokens}    & \makecell{ single-session \\ preference$\uparrow$} & \makecell{single-session \\ assistant $\uparrow$} & \makecell{temporal \\ reasoning $\uparrow$}   & \makecell{multi- \\ session $\uparrow$} & \makecell{knowledge \\ update $\uparrow$} & \makecell{single-session \\ user $\uparrow$}   & overall $\uparrow$ \\ \hline 
MIRIX       & -                 & 53.3                      & 63.6                      & 25.6                  & 30.1          & 52.6              & 72.9                  & 43.49 \\
Zep         & 1.6k              & 53.3                      & \textbf{75.0}                      & 54.1                  & 47.4          & 74.4              & 92.9                  & 63.8 \\
Mem0        & 1.1k              & 90.0                      & 26.8                      & 72.2                  & 63.2          & 66.7              & 82.9                  & 66.4 \\
Memobase    & 1.5k              & 80.1                      & 23.2                      & 75.9                  & 66.9          & \textbf{89.7}     & 92.9                  & 72.4 \\ 
Supermemory & 0.4k              & 89.9                      & 58.9                      & 44.4                  & 52.6          & 55.1              & 85.7                  & 58.4 \\
MemU        & 0.5k              & 76.7                      & 19.6                      & 17.3                  & 42.1          & 41.0              & 67.1                  & 38.4 \\   
\rowcolor[HTML]{EFEFEF} \textsc{MemOS-1031}       & 1.4k              & \textbf{96.7}             & 67.9             & \textbf{77.4}         & \textbf{70.7} & 74.3              & \textbf{95.7}         & \textbf{77.8} \\ \hline
\end{tabular}
}

\label{tab:longmemeval}
\end{table}

%% file: tab/perfeval.tex
\begin{table}[ht]
\centering
\scriptsize
\renewcommand{\arraystretch}{1.2}
\caption{
We evaluated our \textsc{MemOS-1031} model against various baselines on the PreFEval benchmark~\cite{zhao2025llms}, 
all built upon the same GPT-4o-mini\cite{achiam_2023_gpt4} foundation LLM. 
Our comparative analysis involved two scenarios: one without and one with the injection of 10 irrelevant conversation turns. 
We report the Personalized Response percentage (higher is better) and compare it against the percentages of various error types including Preference Unaware, Preference Hallucination, Inconsistency, and Unhelpful Response (lower is better)—along with the context tokens used by each method. 
As the results show, \textsc{MemOS-1031} not only achieved the best Personalized Response performance in both scenarios but also recorded the lowest Preference Unaware error, all while maintaining acceptable error and context length control.
}
\resizebox{\textwidth}{!}{
\begin{tabular}{l|l|ccccc|c}
\hline
                             &Method      & \makecell{Context \\ Token} & \makecell{Preference \\ Unaware$\downarrow$}     & \makecell{Preference \\ Hallucination$\downarrow$} & \makecell{Incon- \\ sistency$\downarrow$} & \makecell{Unhelpful \\ Response$\downarrow$} & \makecell{Personalized \\ Response$\uparrow$}\\ \hline
\multirow{9}{*}{\rotatebox{90}{with 0 turns}}       
                             &Bare LLM        &2027.5        & 80.9                  & \textbf{9.1}               & 0.4           & \textbf{0.0}       & 9.6                     \\
                             &Bare LLM(+rag-5)&393.7          & 19.6                   & 25.9                     & 3.3           & \textbf{0.0}       & 51.2                    \\
                             &MIRIX       &-              & 49.2                  & 9.5                      & \textbf{0.0}   & 3.6                & 37.7                    \\
                             &Mem0        &83.0           & 14.0                  & 18.4                     & 1.4           & 0.3                & 65.9                    \\
                             &Zep         &581.3          & 38.1                  & 18.1                     & 1.7           & 1.4                & 40.7                    \\
                             &Memobase    &430.0          & 36.0                  & 25.8                     & 1.9           & 0.3                & 36.0                    \\
                             &Supermemory &117.0          & 18.5                  & 19.9                     & 2.6           & 0.6                & 58.4                    \\
                             &MemU        &108.1          & 22.1                  & 20.4                     & 2.2           & 1.1                & 54.2                    \\
      & \cellcolor[HTML]{EFEFEF}\textsc{MemOS-1031}  &\cellcolor[HTML]{EFEFEF}557.0          & \cellcolor[HTML]{EFEFEF}\textbf{4.6}         & \cellcolor[HTML]{EFEFEF}14.5                     & \cellcolor[HTML]{EFEFEF}1.6           & \cellcolor[HTML]{EFEFEF}2.1                & \cellcolor[HTML]{EFEFEF}\textbf{77.2}            \\ \hline
\multirow{9}{*}{\rotatebox{90}{with 10 turns}} 
                             &Bare LLM         &11007.5        &93.2                   &\textbf{3.9}           &\textbf{0.1}   &\textbf{0.0}                &2.8  \\
                             &Bare LLM(+rag-5)  &392.5          &26.6                   &27.1                   &3.9            &\textbf{0.0}       &43.2  \\
                             &MIRIX         &-              &77.9                   &72.0                   &\textbf{0.0}   &7.0                &7.9  \\
                             &Mem0          &90.0           &14.8                   &18.4                   &3.1            &\textbf{0.0}       &63.7  \\
                             &Zep           &901.2          &41.0                   &15.7                   &2.1            &1.3                &39.9  \\
                             &Memobase      &563.0          &37.0                   &25.8                   &2.0            &0.1                &34.1  \\
                             &Supermemory   &134.7          &23.9                   &17.2                   &1.8            &0.4                &56.7  \\
                             &MemU          &113.9          &26.5                   &20.3                   &1.1            &0.2                &51.8  \\
      &\cellcolor[HTML]{EFEFEF}\textsc{MemOS-1031}         &\cellcolor[HTML]{EFEFEF}798.7          &\cellcolor[HTML]{EFEFEF}\textbf{7.4}           &\cellcolor[HTML]{EFEFEF}18.6                   &\cellcolor[HTML]{EFEFEF}1.4            &\cellcolor[HTML]{EFEFEF}0.7                &\cellcolor[HTML]{EFEFEF}\textbf{71.9}  \\ \hline
\end{tabular}
}

\label{tab:perfeval}
\end{table}

%% file: tab/personamem.tex
\begin{table}[ht]
\centering
\scriptsize
\renewcommand{\arraystretch}{1.1}
\caption{
We evaluated our \textsc{MemOS-1031} model against various baselines on the PersonaMem~\cite{jiang2025know} benchmark, 
all built upon the same GPT-4o-mini\cite{achiam_2023_gpt4} foundation LLM. 
We report the precision of the chosen one-in-four along with the context tokens used by each method. 
As the results show, \textsc{MemOS-1031} achieved the best precision while maintaining acceptable context length control.
}
\resizebox{\textwidth}{!}{
\begin{tabular}{l|cccccc>{\columncolor[HTML]{EFEFEF}}c}
\hline
Method          & MIRIX     & Mem0  & Zep   & Memobase  &  MemU & Supermemory   & \textsc{MemOS-1031} \\ \hline
Precision(1 in 4)$\uparrow$       & 38.4      & 43.1  & 57.8  & 58.9      & 56.8  & 53.9          & \textbf{61.2} \\
context tokens  &  -        & 140   & 1657  & 2092      & 496   & 204           & 1424 \\ \hline
\end{tabular}
}
\label{tab:personamem}
\end{table}

%% file: tab/latency.tex
\begin{table}[ht]
\centering
\scriptsize
\renewcommand{\arraystretch}{1.4}
\caption{
We evaluated the latency and LLM evaluation scores of various API methods on the LoCoMo benchmark under different QPS (queries per second) pressure of network request. 
Metrics reported include the P99, P90, and mean latency, as well as the success percentage rate for both add and search operations. 
Our results show that \textsc{MemOS-1031} exhibited the most robustness, maintaining the lowest latency across nearly all metrics regardless of the QPS pressure. 
Furthermore, \textsc{MemOS-1031} achieved a 100\% success rate and maintained reasonable latency even under 100 QPS.
}
\resizebox{\textwidth}{!}{
\begin{tabular}{l|c|cccc|cccc}
\hline
                        &               & \multicolumn{4}{c|}{Add}                                                                                          & \multicolumn{4}{c}{Search} \\ \cline{3-10}
QPS                     & API           & P99(ms) $\downarrow$  & P90(ms) $\downarrow$  & \makecell{mean \\ latency \\(ms) $\downarrow$} & \makecell{success \\ percentage \\ (\%) $\uparrow$} & P99(ms) $\downarrow$  & P90(ms) $\downarrow$  & \makecell{mean \\ latency \\(ms) $\downarrow$} & \makecell{success \\ percentage \\ (\%) $\uparrow$}  \\\hline
\multirow{6}{*}{10 qps} & Mem0          & 2841.0                & 1650.0                & 888.0                         & \textbf{100.0}                    & 4637.0                & 1397.0                & 1089.0                        & \textbf{100.0} \\
                        & Memobase      & 6169.1                & 1533.1                & 1126.2                        & 99.9                              & 14856.7               & 5974.7                & 2105.2                        & 99.9  \\
                        & Supermemory   & 3467.4                & 1816.9                & 1446.3                        & \textbf{100.0}                    & 2107.6                & 1457.4                & 1057.1                        & \textbf{100.0} \\
                        & MemU          & 12070.6               & 7273.1                & 5077.9                        & 64.2                              & 63000.7               & 60539.5               & 47554.5                       & 7.7   \\
                        & Zep           & \textbf{375.1}        & 254.5                 & 239.0                         & 99.7                              & 5348.7                & 614.8                 & 571.7                         & 99.9  \\
                        & \cellcolor[HTML]{EFEFEF}\textsc{MemOS-1031}    & \cellcolor[HTML]{EFEFEF}376.4                 & \cellcolor[HTML]{EFEFEF}\textbf{211.6}        & \cellcolor[HTML]{EFEFEF}\textbf{191.9}                & \cellcolor[HTML]{EFEFEF}\textbf{100.0}                    & \cellcolor[HTML]{EFEFEF}\textbf{777.1}        & \cellcolor[HTML]{EFEFEF}\textbf{528.4}        & \cellcolor[HTML]{EFEFEF}\textbf{440.5}                & \cellcolor[HTML]{EFEFEF}\textbf{100.0}   \\ \hline
\multirow{6}{*}{40 qps} & Mem0          & 2624.0                & 1258.0                & 672.89                        & 41.7                              & 1723.0                & 1309.0                & 698.22                        & 51.2  \\
                        & Memobase      & 7074.0                & 6251.5                & 2794.4                        & 68.8                              & 39985.3               & 18822.0               & 5515.5                        & 97.8  \\
                        & Supermemory   & 60639.1               & 50951.5               & 24996.6                       & 89.4                              & 2050.8                & 1117.2                & 926.0                         & \textbf{100.0} \\
                        & MemU          & 42129.9               & 35730.3               & 29033.0                       & 7.5                               & 61944.1               & 60565.2               & 41042.7                       & 3.8   \\
                        & Zep           & \textbf{269.9}        & 240.5                 & 227.7                         & 26.6                              & 5802.5                & 1480.2                & 670.0                         & 26.8  \\
                        & \cellcolor[HTML]{EFEFEF}\textsc{MemOS-1031}    & \cellcolor[HTML]{EFEFEF}282.1                 & \cellcolor[HTML]{EFEFEF}\textbf{222}          & \cellcolor[HTML]{EFEFEF}\textbf{192.6}                & \cellcolor[HTML]{EFEFEF}\textbf{100.0}                      & \cellcolor[HTML]{EFEFEF}\textbf{951.6}        & \cellcolor[HTML]{EFEFEF}\textbf{767.6}        & \cellcolor[HTML]{EFEFEF}\textbf{613.8}                & \cellcolor[HTML]{EFEFEF}\textbf{100.0} \\ \hline
100 qps                 & \cellcolor[HTML]{EFEFEF}\textsc{MemOS-1031}    & \cellcolor[HTML]{EFEFEF}463.2                 & \cellcolor[HTML]{EFEFEF}311.1                 & \cellcolor[HTML]{EFEFEF}251.9                         & \cellcolor[HTML]{EFEFEF}100.0                              & \cellcolor[HTML]{EFEFEF}1171.1                & \cellcolor[HTML]{EFEFEF}872.5                 & \cellcolor[HTML]{EFEFEF}741.2                         & \cellcolor[HTML]{EFEFEF}100.0 \\ \hline
\end{tabular}
}

\label{tab:time_latency}
\end{table}

%% file: tab/kvtest.tex
\begin{table}[ht]
\centering
\scriptsize
\renewcommand{\arraystretch}{1.1}
\caption{
Evaluation of Time to First Token (TTFT) and acceleration effect across different models, context lengths, and query lengths using the HuggingFace \texttt{transformers} library. 
We compare two memory injection strategies: direct prompt-based injection and KV-based attention cache injection.
Gray-highlighted rows correspond to \textsc{MemOS}’s strategy, which uses KV-form memory injection and consistently achieves faster response without altering output semantics.
}
\resizebox{\textwidth}{!}{%
\begin{tabular}{l|lc|lc|c
>{\columncolor[HTML]{EFEFEF}}c c
>{\columncolor[HTML]{EFEFEF}}c }
\hline
Model                         & Ctx                      & CtxTok                 & Qry    & QryTok & Build (s) & KV TTFT (s) & Dir TTFT (s) & Speedup (\%) \\ \hline
                              &                          &                        & long   & 952.7  & 0.92      & 0.50        & 2.37         & 79.1         \\
                              &                          &                        & medium & 302.7  & 0.93      & 0.19        & 2.16         & 91.1         \\
                              & \multirow{-3}{*}{long}   & \multirow{-3}{*}{6064} & short  & 167    & 0.93      & 0.12        & 2.04         & 94.2         \\ \cline{2-9} 
                              &                          &                        & long   & 952.7  & 0.41      & 0.43        & 1.22         & 64.6         \\
                              &                          &                        & medium & 302.7  & 0.41      & 0.16        & 1.08         & 85.1         \\
                              & \multirow{-3}{*}{medium} & \multirow{-3}{*}{2773} & short  & 167    & 0.43      & 0.10        & 0.95         & 89.7         \\ \cline{2-9} 
                              &                          &                        & long   & 952.7  & 0.12      & 0.39        & 0.51         & 23.0         \\
                              &                          &                        & medium & 302.7  & 0.12      & 0.14        & 0.32         & 55.6         \\
\multirow{-9}{*}{Qwen3-8B}    & \multirow{-3}{*}{short}  & \multirow{-3}{*}{583}  & short  & 167    & 0.12      & 0.08        & 0.29         & 71.3         \\ \hline
                              &                          &                        & long   & 952.7  & 0.71      & 0.31        & 1.09         & 71.4         \\
                              &                          &                        & medium & 302.7  & 0.71      & 0.15        & 0.98         & 84.3         \\
                              & \multirow{-3}{*}{long}   & \multirow{-3}{*}{6064} & short  & 167    & 0.71      & 0.11        & 0.96         & 88.8         \\ \cline{2-9} 
                              &                          &                        & long   & 952.7  & 0.31      & 0.24        & 0.56         & 56.9         \\
                              &                          &                        & medium & 302.7  & 0.31      & 0.12        & 0.47         & 75.1         \\
                              & \multirow{-3}{*}{medium} & \multirow{-3}{*}{2773} & short  & 167    & 0.31      & 0.08        & 0.44         & 81.2         \\ \cline{2-9} 
                              &                          &                        & long   & 952.7  & 0.09      & 0.20        & 0.24         & 18.6         \\
                              &                          &                        & medium & 302.7  & 0.09      & 0.09        & 0.15         & 39.6         \\
\multirow{-9}{*}{Qwen3-32B}   & \multirow{-3}{*}{short}  & \multirow{-3}{*}{583}  & short  & 167    & 0.09      & 0.07        & 0.14         & 53.5         \\ \hline
                              &                          &                        & long   & 952.7  & 1.26      & 0.48        & 2.04         & 76.4         \\
                              &                          &                        & medium & 302.7  & 1.26      & 0.23        & 1.82         & 87.2         \\
                              & \multirow{-3}{*}{long}   & \multirow{-3}{*}{6064} & short  & 167    & 1.27      & 0.15        & 1.79         & 91.4         \\ \cline{2-9} 
                              &                          &                        & long   & 952.7  & 0.58      & 0.39        & 1.05         & 62.7         \\
                              &                          &                        & medium & 302.7  & 0.58      & 0.18        & 0.89         & 79.2         \\
                              & \multirow{-3}{*}{medium} & \multirow{-3}{*}{2773} & short  & 167    & 0.71      & 0.23        & 0.82         & 71.6         \\ \cline{2-9} 
                              &                          &                        & long   & 952.7  & 0.16      & 0.33        & 0.43         & 23.8         \\
                              &                          &                        & medium & 302.7  & 0.16      & 0.15        & 0.27         & 43.2         \\
\multirow{-9}{*}{Qwen2.5-72B} & \multirow{-3}{*}{short}  & \multirow{-3}{*}{583}  & short  & 167    & 0.16      & 0.10        & 0.25         & 60.5         \\ \hline
\end{tabular}
}
\label{tab:ttft-performance}
\end{table}